\documentclass[acmtog]{acmart}
\AtBeginDocument{%
  \providecommand\BibTeX{{%
    \normalfont B\kern-0.5em{\scshape i\kern-0.25em b}\kern-0.8em\TeX}}}

\newcommand{\MYhref}[3][blue]{\href{#2}{\color{#1}{#3}}}%

\acmSubmissionID{260}

\begin{document}

\title{DiFT: Differentiable Differential Feature Transform for Multi-View Stereo}
\author{Kaizhang Kang}
\author{Chong Zeng}
\author{Hongzhi Wu$^*$}\thanks{*: corresponding author (\MYhref{mailto:hwu@acm.org}{hwu@acm.org}).}
\affiliation{%
  \institution{State Key Lab of CAD\&CG, Zhejiang University}
  \country{China}
}

\author{Kun Zhou}
\affiliation{%
  \institution{State Key Lab of CAD\&CG, Zhejiang University and ZJU-FaceUnity Joint Lab of Intelligent Graphics}
  \country{China}
}

\renewcommand{\shortauthors}{Trovato and Tobin, et al.}

\newcommand{\figref}[1]{Fig.~\ref{#1}}
\newcommand{\tabref}[1]{Tab.~\ref{#1}}
\newcommand{\appref}[1]{Appendix~\ref{#1}}
\newcommand{\stepref}[1]{Step~\ref{#1}}
\newcommand{\eqnref}[1]{Eq.~\ref{#1}}
\newcommand{\algref}[1]{Algorithm~\ref{#1}}
\newcommand{\note}[1]{{\color{blue} #1}}
\newcommand{\warning}[1]{{\color{red} #1}}
\def\sec#1{Sec.~\ref{#1}}

\begin{abstract}
We present a novel framework to automatically learn to transform the differential cues from a stack of images densely captured with a rotational motion into spatially discriminative and view-invariant per-pixel features at each view. These low-level features can be directly fed to any existing multi-view stereo technique for enhanced 3D reconstruction. The lighting condition during acquisition can also be jointly optimized in a differentiable fashion. We sample from a dozen of pre-scanned objects with a wide variety of geometry and reflectance to synthesize a large amount of high-quality training data. The effectiveness of our features is demonstrated on a number of challenging objects acquired with a lightstage, comparing favorably with state-of-the-art techniques. Finally, we explore additional applications of geometric detail visualization and computational stylization of complex appearance.
\end{abstract}


\begin{teaserfigure}
  \includegraphics[height = 1.4in]{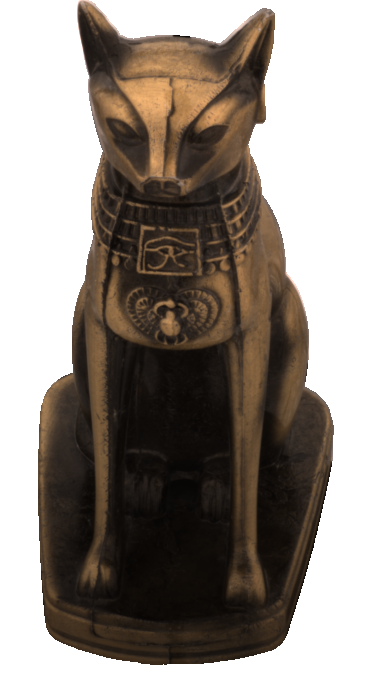}
  \includegraphics[height = 1.4in]{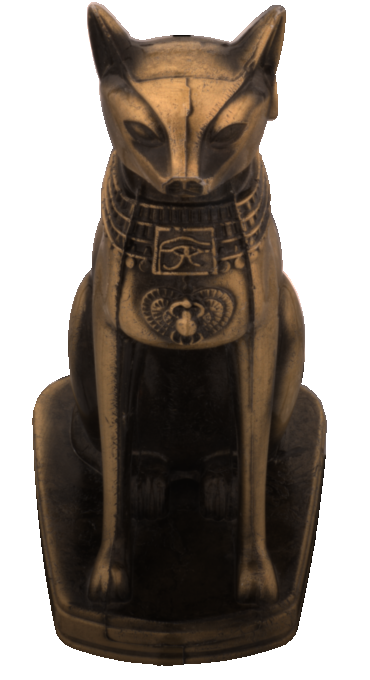}
  \includegraphics[height = 1.4in]{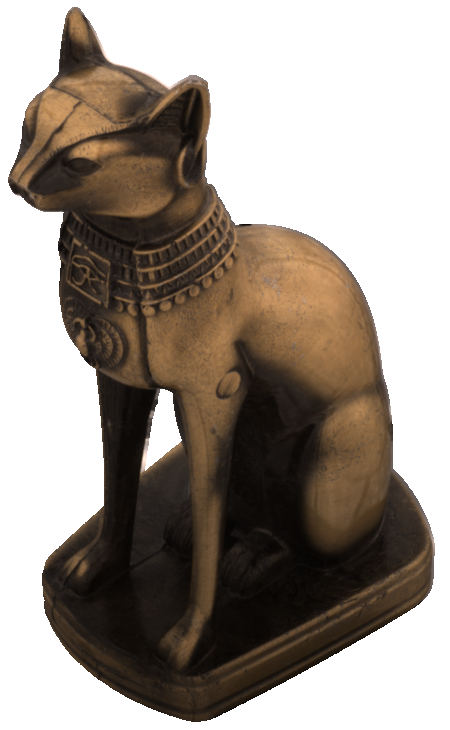}
  \includegraphics[height = 1.4in]{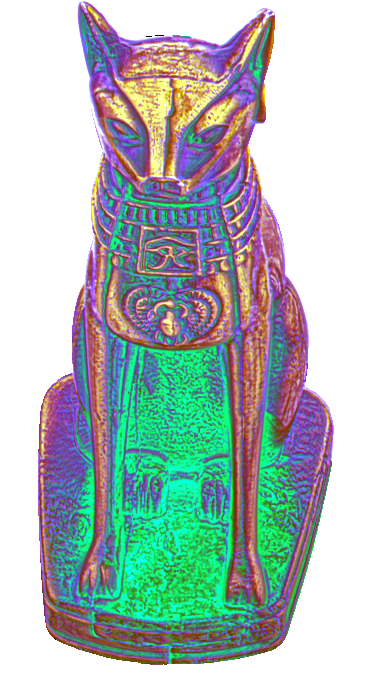}
  \includegraphics[height = 1.4in]{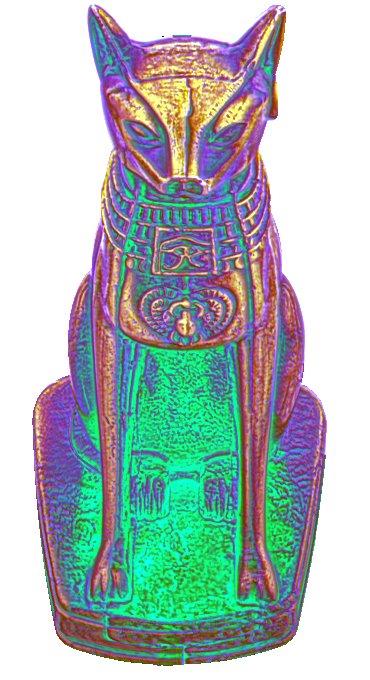}
  \includegraphics[height = 1.4in]{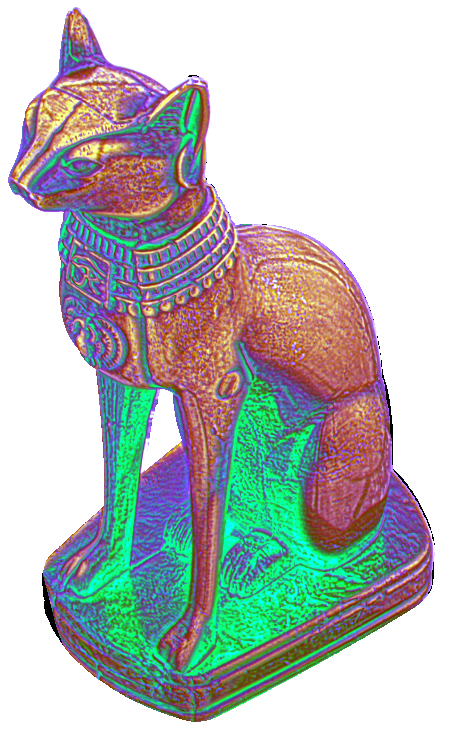}
  \includegraphics[height = 1.4in]{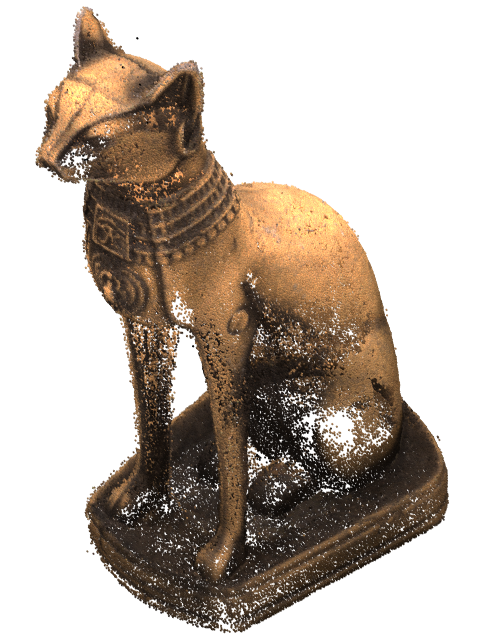}
  \includegraphics[height = 1.4in]{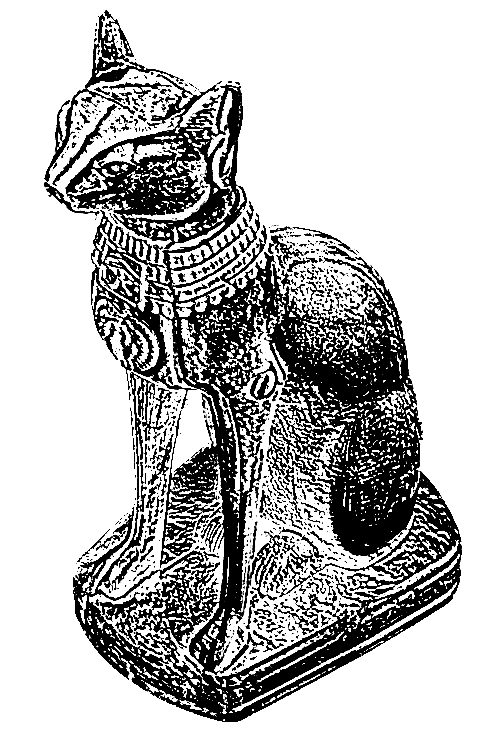}
  \caption{From input images densely captured with a rotational motion under pre-optimized illumination (left), we automatically learn to transform the differential structural cues into spatially discriminative and view-invariant per-pixel features at each view (center), which can be directly fed to any multi-view stereo technique for enhanced 3D reconstruction of an object with challenging appearance (second to the right). We also explore additional applications like computational stylization (right).}
  \label{fig:teaser}
\end{teaserfigure}

\maketitle

\section{Introduction}

Image-based 3D reconstruction of physical objects with complex appearance is a key problem in computer vision and graphics. It has many important applications, ranging from quality inspection, product design, e-commerce to cultural heritage,.

It is known that images captured with densely sampled views exhibit unique \textbf{differential} structural cues that can be used to deduce more accurate and complete 3D information~\cite{SURVEY:2017}, in comparison with standard multi-view stereo methods~\cite{furukawa2015multi} that do not explicitly consider such properties. For example, the trace of an unoccluded 3D point is a straight line in a stack of images captured with an equidistant 1D translational motion between the camera and the object; and the slope of this line directly corresponds to the depth. In the case of a rotational motion, the trace of a point is a helix, whose radius and pitch can be transformed to obtain a 3D position.

However, it remains an open problem to identify as many traces and compute their differentiable properties as accurately as possible for 3D reconstruction of a general object. Specifically, it is highly challenging to handle the appearance that varies considerably with the lighting and/or view conditions, textureless regions and complex occlusions. Dense depth estimation from a lightfield introduces heuristic~\cite{wanner2013variational} or learning-based~\cite{heber2016convolutional} solutions, often assuming a Lambertian reflectance. Differential photometric stereo manually derives the relationship between depths and derivatives of image pairs undergoing differential motions~\cite{Chandraker:2013:DIFF,wang2016svbrdf}. The closest work to ours is~\cite{Kang:2021:LEARN}, which learns to convert the photometric information into low-level multi-view stereo features. Their approach relies on efficient illumination sampling, and does not exploit differential cues in the view domain.

In this paper, we present a novel \textbf{differentiable} framework to automatically learn to extract the geometric information from differential cues in a stack of images captured with an equiangular rotational motion, and transform into spatially distinctive and view-invariant per-pixel features at each view, in an end-to-end fashion. The illumination condition during acquisition can also be jointly optimized with the feature transform. We focus on learning high-quality local features, and delegate subsequent processing (e.g., spatial aggregation/view selection) to any existing learning-/non-learning-based multi-view stereo pipeline. In addition, our framework is flexible, as it can adapt to various factors, including the type of motion and the lighting layout, in a data-driven manner.




The effectiveness of the framework is demonstrated with a lightstage, on a variety of challenging objects with complex appearance. Our reconstruction results compare favorably with state-of-the-art techniques. Moreover, we perform an extensive study on the impact of different factors over the reconstruction quality, and extend the framework to a fixed illumination condition. Finally, we explore additional applications of visualization of minor geometric details as well as computational stylization of high-dimensional appearance.

\section{Related Work}


\subsection{Depth from Densely Sampled Views}

A straightforward solution is to treat the problem as standard multi-view stereo~\cite{Vaish:2006:SA,bishop2009light}. However, it does not explicitly exploit the rich, depth-related structures in densely sampled near-by views, leading to sub-optimal geometric reconstruction~\cite{SURVEY:2017}. In addition, the narrow baseline between neighboring views may require special treatments~\cite{jeon2015accurate}.

From the seminal work of~\cite{bolles1987epipolar}, substantial research efforts have been devoted to dense depth estimation from a lightfield that densely samples the view domain~\cite{wu2017light}. A typical pipeline consists of three steps. The first step extracts useful local structural cues~\cite{wanner2013variational,zhang2016robust} from the input data (e.g., EPIs or virtual views sampled on circles~\cite{heber2013variational}). Here Lambertian reflectance is 
usually assumed. Additional care must be taken to handle view-varying appearance variations, where it is difficult to compute reliable local features~\cite{tao2014depth,sulc2022recovery}. The next step aggregates local information to compute a global representation. Sophisticated algorithms are proposed to handle challenging cases like complex occlusions~\cite{chen2014light,wang2016depth}. The final step is to generate the depth according to its relationship with the global information~\cite{johannsen2016sparse,wanner2012globally}. Recently, deep learning is adopted to convert the input data to depth-related orientation cues~\cite{heber2016convolutional}, or even to a disparity map in an end-to-end fashion~\cite{shin2018epinet}.

In differential photometric stereo~\cite{Chandraker:2013:DIFF,wang2016svbrdf}, the relationship among the depth, the normal and derivatives of image pairs undergoing differential motions in the presence of diffuse plus single-lobe SVBRDFs are manually analyzed. Building upon this relationship, efficient methods are proposed to optimize for a depth map.

In comparison, our framework is the first to automatically learn efficient, low-level geometric features from the 3D image stack captured with densely sampled views, in the presence of complex appearance. Existing work either is based on manual derivations, or only works well with Lambertian reflectance. Moreover, we jointly optimize active illumination condition during acquisition to pack more geometric information into physical measurements, essentially improving the signal-to-noise-ratio. Third, we focus on learning modular features and delegate local-to-global processing to powerful multi-view stereo techniques, making it possible to harness the recent or even future development in that field.

\subsection{Features for Multi-View Stereo}
A typical pipeline in multi-view stereo~\cite{furukawa2015multi} starts with computing local features from each input image. It then matches the features across different views, and exploits their correspondences to compute 3D points via triangulation. The quality of features directly affects the number and the reliability of correspondence matches, which is critical for the completeness and accuracy of the reconstructed shape.

Over the past decades, the research on feature design has gradually
evolved from hand-crafted ones~\cite{mikolajczyk2005performance,fan2015local}, traditional-learning-based ones~\cite{ke2004pca,trzcinski2013boosting,simonyan2014learning} to deep-learning-based ones~\cite{zbontar2015computing,zagoruyko2015learning,tian2017l2}. However, the majority of work assumes Lambertian reflectance and attempts to filter out, rather than exploit, the complex appearance variations.

The closest work to ours is~\cite{Kang:2021:LEARN}, which learns to convert the photometric information into spatially discriminative and view-invariant low-level features that are amenable for multi-view stereo. Their approach requires sufficient photometric information, and does not work as good as ours in the case of a single lighting condition. Moreover, it does not exploit the rich differential cues available in densely sampled views. 

\section{Acquisition Device}
We conduct physical experiments in a box-shaped lightstage with a size of 80cm$\times$80cm$\times$80cm, as illustrated in the left of~\figref{fig:pipeline}. A 24MP Basler a2A5328-15ucPRO vision camera takes photographs of a 3D object placed on a digital turntable in the lightstage, from an angle of about $45^{\circ}$ above the horizontal plane. The sample object is illuminated with 24,576 high-intensity LEDs mounted to all six faces of the box. The LED pitch is 1cm, and the intensity is quantized with 8 bits and controlled with house-made circuits. We calibrate the intrinsic and extrinsic parameters of the camera, as well as the positions, orientations and the angular intensity distribution of each LED.

\section{Rendering Equation}
The following equation describes the relationship among the image measurement $B$ from a surface point $\mathbf{p}$, the reflectance $f$ and the intensity $I$ of each LED of the device, which is crucial for optimization in a differentiable framework. Here we focus on a single channel for brevity.
\begin{align}
B(I, \mathbf{x_{p}}, \mathbf{n_{p}}, \mathbf{t_{p}}, f) =  &\sum_{l}  I(l)  \int  \frac{1}{|| \mathbf{x_{l}} - \mathbf{x_{p}} ||^2} \Psi(\mathbf{x_{l}}, -\mathbf{\omega_{i}}) V(\mathbf{x_{l}}, \mathbf{x_{p}})  \nonumber \\
& f(\mathbf{\omega_{i}}'; \mathbf{\omega_{o}}', \mathbf{p}) (\mathbf{\omega_{i}} \cdot \mathbf{n_{p}})^{+} (-\mathbf{\omega_{i}} \cdot \mathbf{n_{l}})^{+} d\mathbf{x_{l}},
\label{eq:render}
\end{align}
where $l$ is the index of a planar light source, and $I(l)$ is its intensity in the range of [0, 1], the collection of which will be referred to as a \textbf{lighting pattern}. In addition, $\mathbf{x_{p}}$/$\mathbf{n_{p}}$/$\mathbf{t_{p}}$ is the position/normal/tangent of $\mathbf{p}$, while $\mathbf{x_{l}}$/$\mathbf{n_{l}}$ is the position/normal of a point on the light whose index is $l$. We denote $\mathbf{\omega_{i}}$/$\mathbf{\omega_{o}}$ as the lighting/view direction, with $\mathbf{\omega_{i}} = \frac{\mathbf{x_{l}} - \mathbf{x_{p}}}{|| \mathbf{x_{l}} - \mathbf{x_{p}} ||}$. $\Psi(\mathbf{x_{l}}, \cdot)$ is the angular distribution of the light intensity. $V$ is a binary visibility function between $\mathbf{x_{l}}$ and $\mathbf{x_{p}}$. The operator $( \cdot )^{+}$ computes the dot product between two vectors, and clamps a negative result to zero. $f(\cdot;  \mathbf{\omega_{o}}', \mathbf{p})$ is a 2D BRDF slice, which is a function of the lighting direction. We use the anisotropic GGX model~\cite{Walter:2007:GGX} to represent $f$. 


\section{Overview}
 \begin{figure*}
     \includegraphics[width = \linewidth]{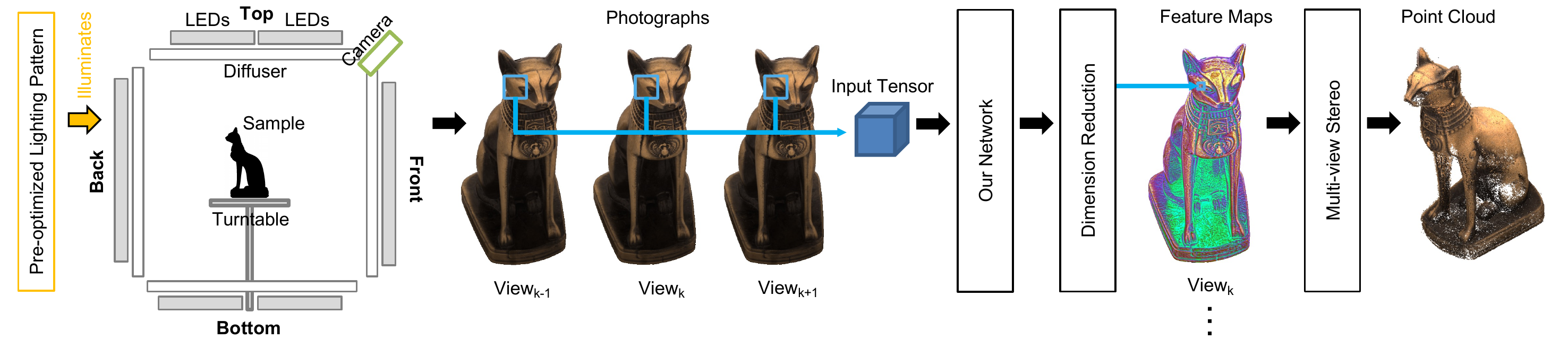}
   \caption{Our pipeline. We first capture a series of images under a pre-optimized lighting pattern using a lightstage, with the object rotating from 0 to $360^{\circ}$.  For each valid pixel at each view, our network transforms the corresponding local neighborhood in the spatial-angular domain to a spatially discriminative and view/lighting-invariant low-level feature vector, resulting in multi-view feature maps. Finally, the maps are converted to RGB images and sent to a state-of-the-art multi-view stereo technique to reconstruct a 3D shape. Only 3 views are shown here for illustration purposes.}
   \label{fig:pipeline}
 \end{figure*}
We first acquire a series of images of a physical object, rotating from 0 to $360^{\circ}$ with a constant angular interval under a pre-optimized lighting pattern. From high-quality training data synthesized with a wide variety of pre-captured objects, our network automatically learns to transform a local rank-3 tensor in the \textbf{spatial-angular} domain (i.e., the differential structural cues) to a spatially discriminative and view/lighting-invariant low-level feature vector. The procedure is repeated for each pixel in each image, resulting in multi-view high-dimensional feature maps. Finally, the maps are converted to RGB images and sent to a state-of-the-art multi-view stereo technique to reconstruct a 3D shape. \figref{fig:pipeline} illustrates the process.

\section{Our Network}

\subsection{Input/Output}
\label{sec:inputoutput}

For each valid pixel at each input view, we assemble its neighborhood in the spatial-angular domain into a rank-3 tensor of $w_s \times w_s \times w_a$, with the pixel of interest sitting at the center ($w_s = w_a = 5$ in most experiments). This tensor is the main input to our network. Similar to~\cite{Kang:2021:LEARN}, the view specification of the current image is also fed to the network in the continuous form of [$\cos(\theta)$, $\sin(\theta)$], where $\theta$ is the absolute rotation angle of the turntable in our setup. The idea is to encourage the network to exploit this additional view information to enhance feature quality. Our output is a 10D unit vector, representing a spatially distinctive and view-invariant feature at the pixel of interest.

\subsection{Architecture}
\begin{figure}
    \includegraphics[width = \columnwidth]{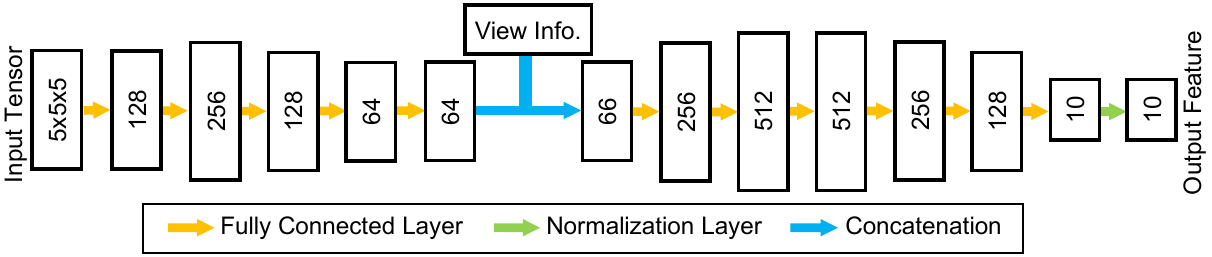}
  \caption{Network architecture. Our network takes as input a rank-3 tensor and produces a normalized 10D feature as output.}
  \label{fig:net}
\end{figure}
Our main network is designed to exploit the differential structural cues in the input tensor. It consists of 11 fc layers, 1 normalization layer and employs leaky ReLU for nonlinear activation. The first layer can be viewed as filters that extract different structural cues from the input tensor (\figref{fig:filter}). Note that the view specification (\sec{sec:inputoutput}) is supplied to the network after 5 layers. The idea is to perform a low-level, view-independent transform first, and then further convert into a view-dependent feature with the additional view information. The final normalization layer produces a unit feature vector as output for training stability, as common in related literature~\cite{schroff2015facenet,wu2017sampling}. Please refer to~\figref{fig:net} for an illustration.

In addition, the lighting pattern during acquisition is related to the input tensor (and therefore the loss function) in a differentiable way, according to~\eqnref{eq:render}. This allows us to optimize the illumination in conjunction with our main network for further quality improvement, as detailed in~\sec{sec:data}. To constrain the intensity of each light to the feasible range of [0,1], we use two unconstrained parameters $a$ and $b$ to express one LED intensity as $\frac{1}{2} (\frac{a}{\sqrt{a^2+b^2}} + 1)$.

\subsection{Data Generation}
\label{sec:data}
\begin{figure}
    \includegraphics[width = 0.19\linewidth]{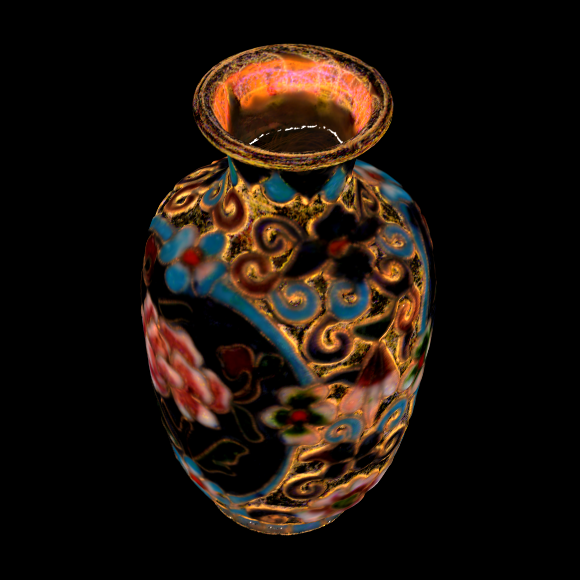}
    \includegraphics[width = 0.19\linewidth]{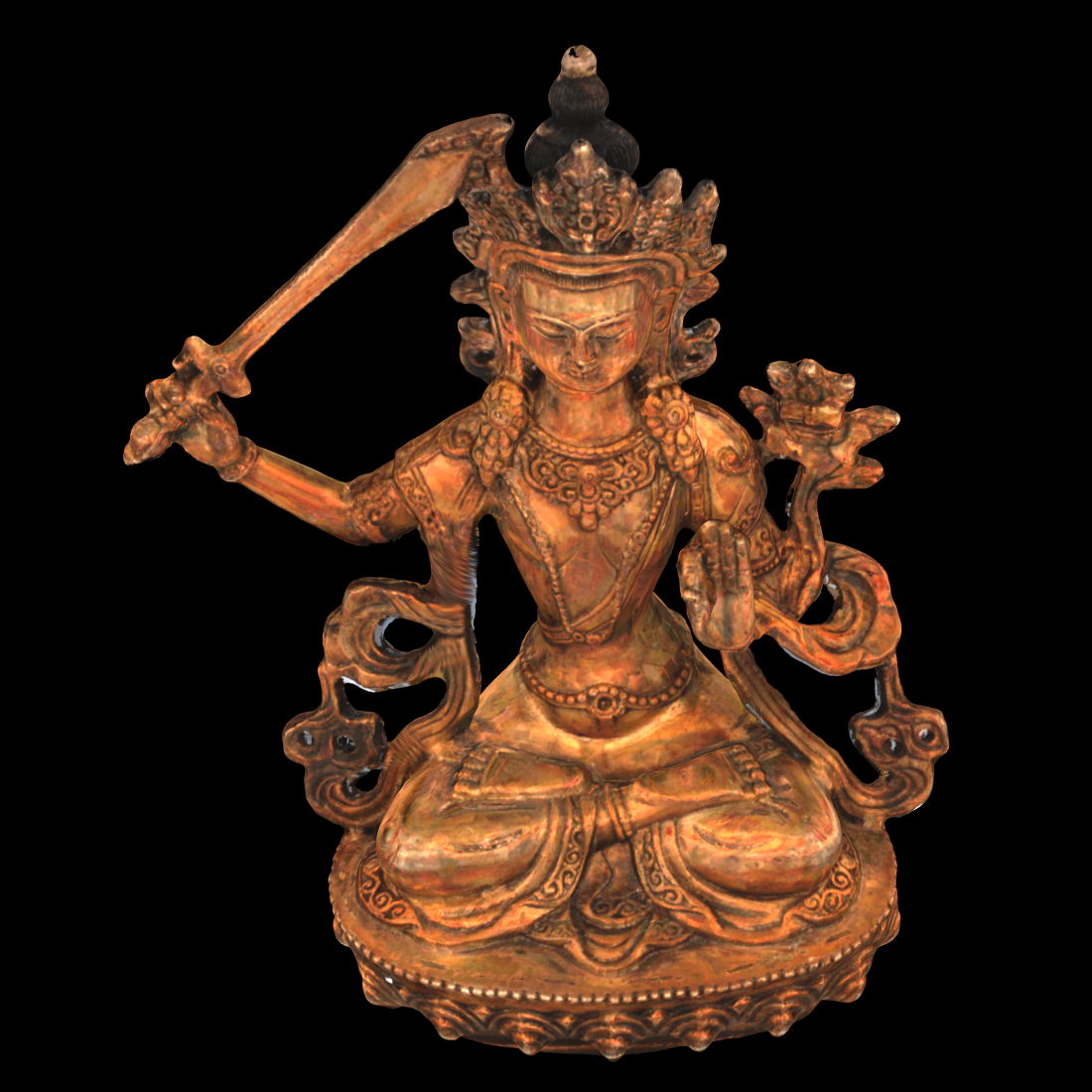}
    \includegraphics[width = 0.19\linewidth]{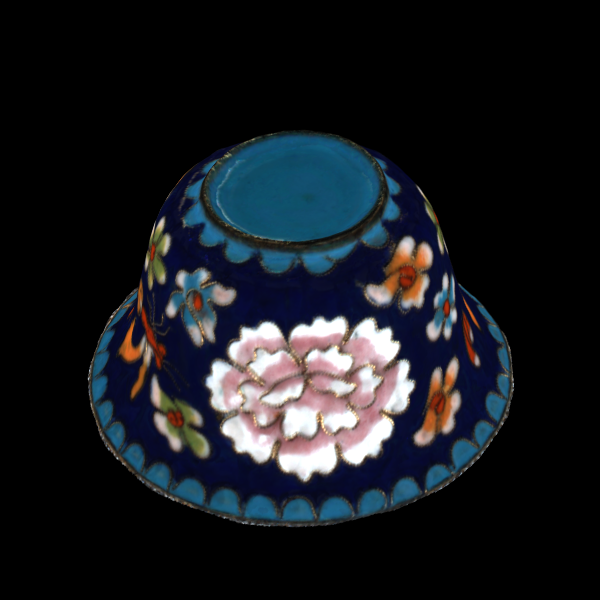}
    \includegraphics[width = 0.19\linewidth]{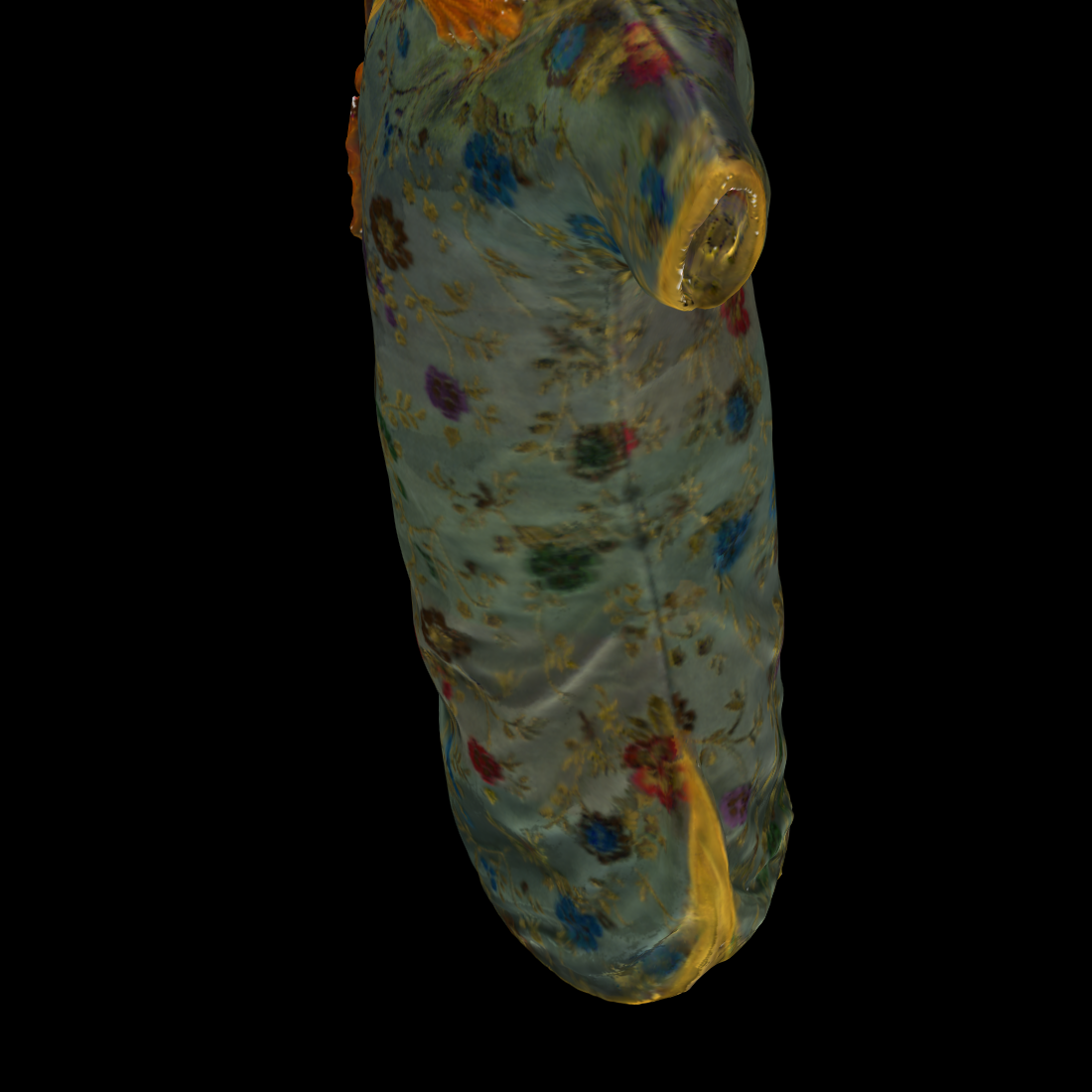}
    \includegraphics[width = 0.19\linewidth]{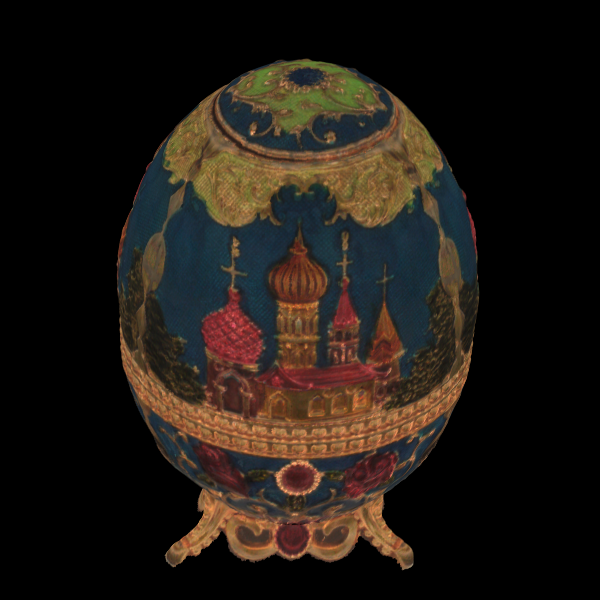}
    \includegraphics[width = 0.19\linewidth]{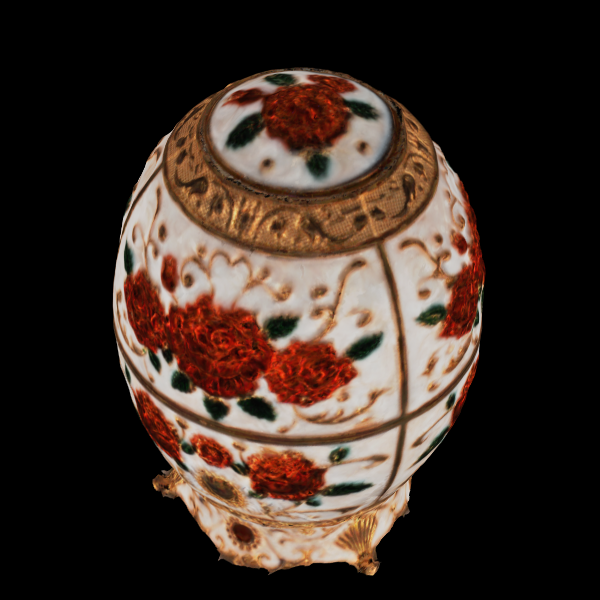}
    \includegraphics[width = 0.19\linewidth]{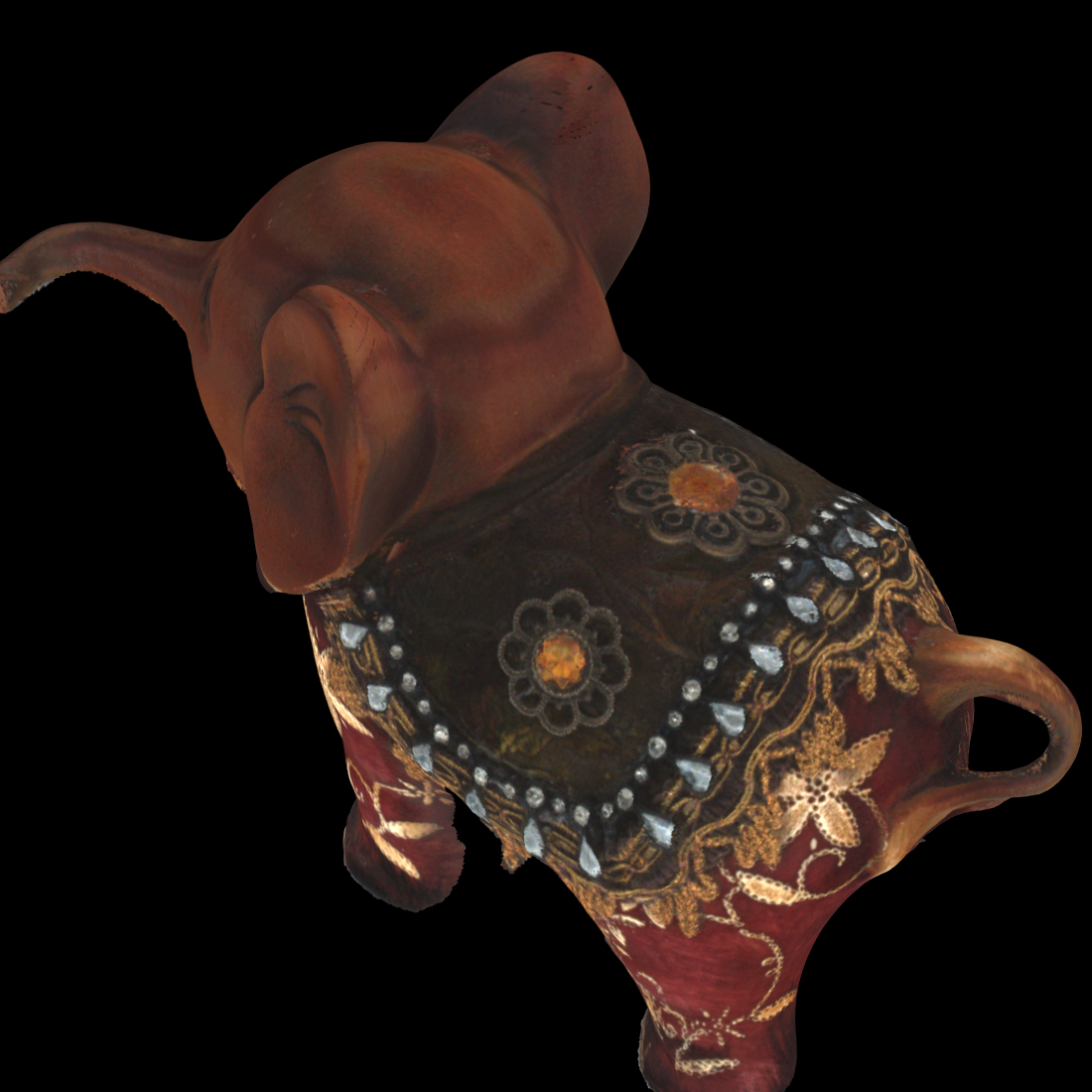}
    \includegraphics[width = 0.19\linewidth]{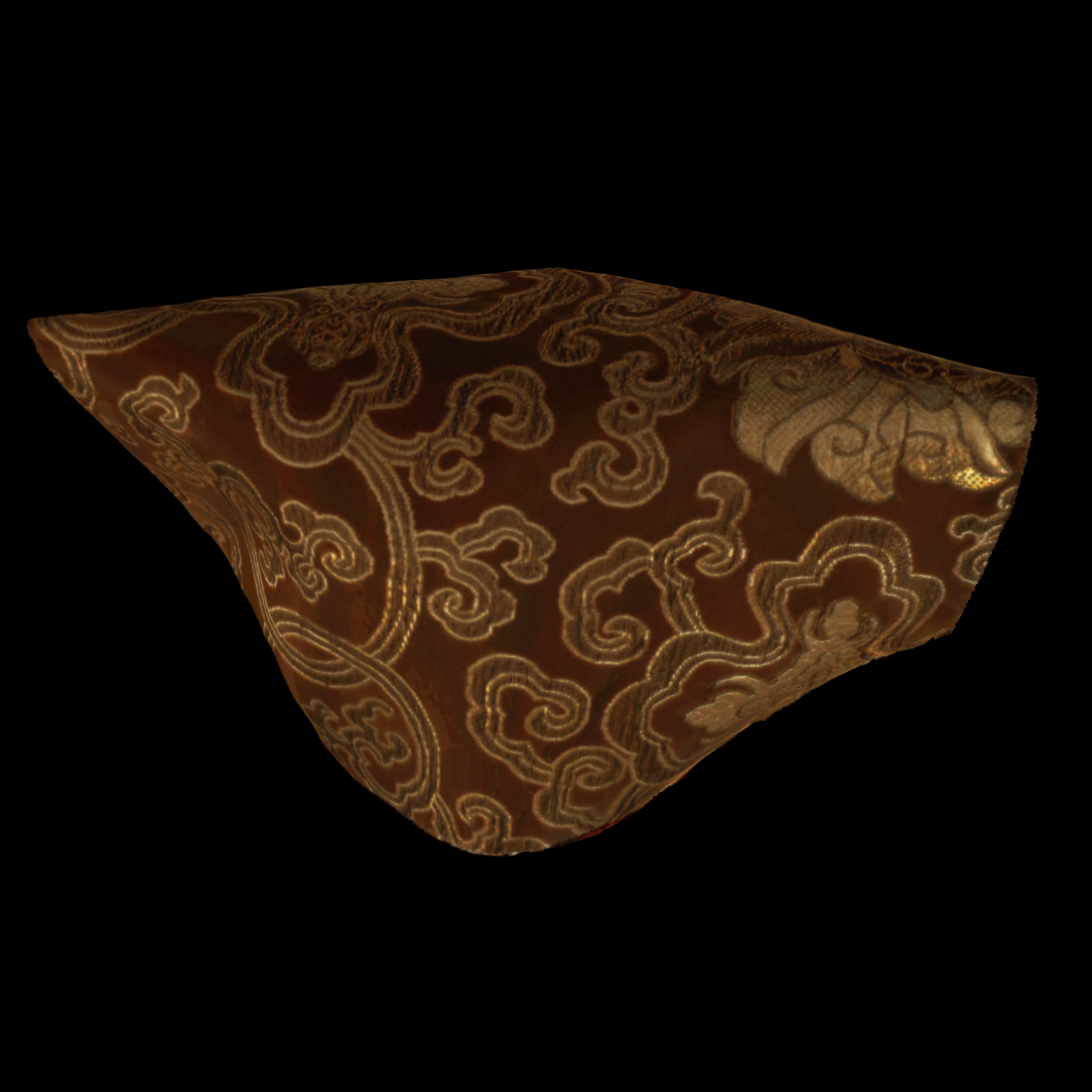}
    \includegraphics[width = 0.19\linewidth]{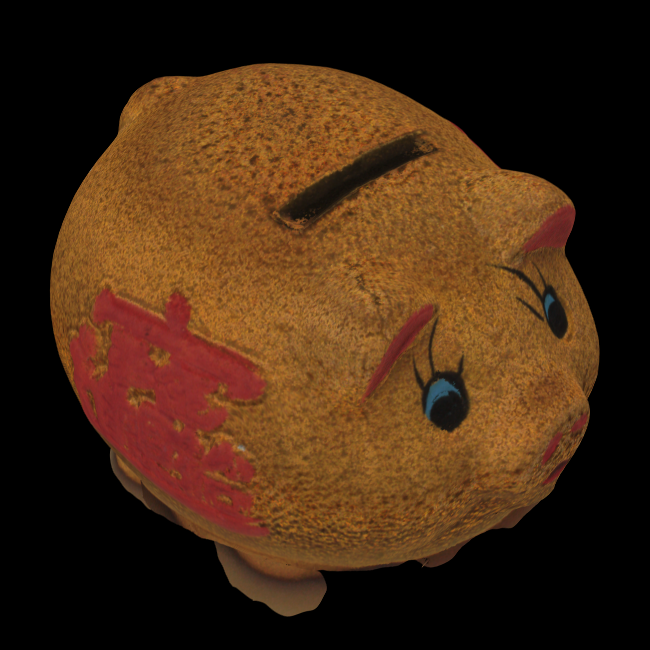}
    \includegraphics[width = 0.19\linewidth]{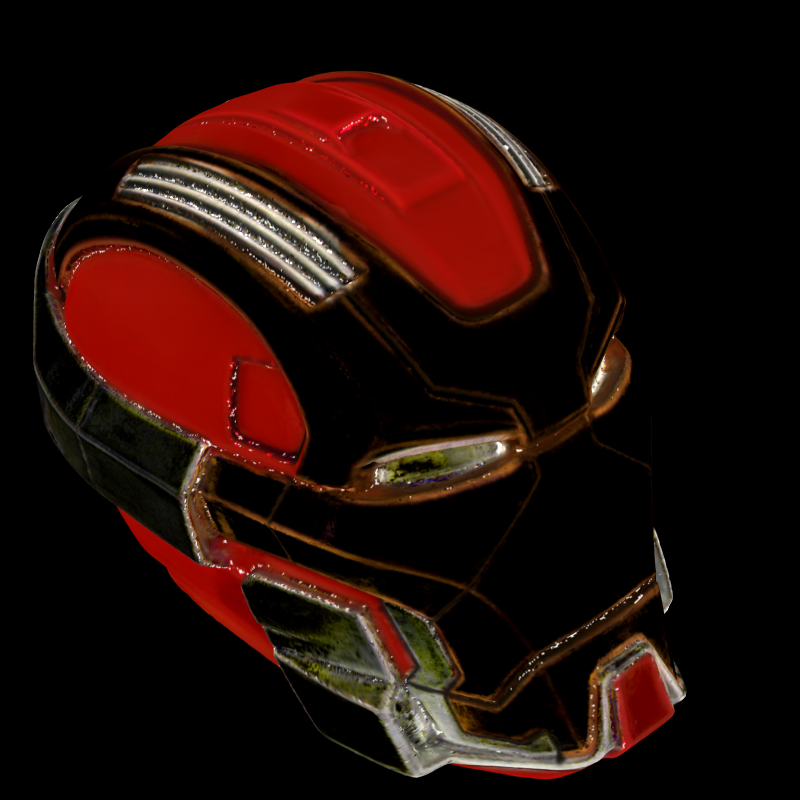}
    \includegraphics[width = 0.19\linewidth]{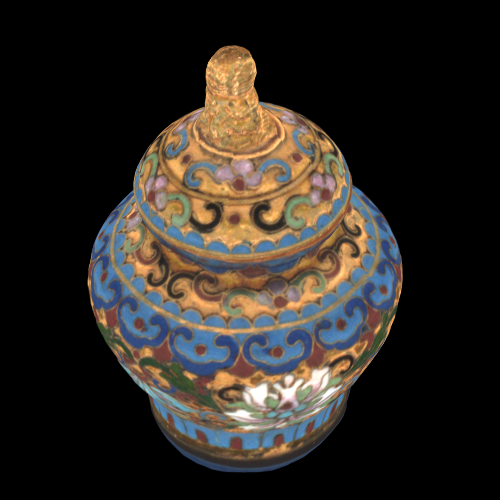}
    \includegraphics[width = 0.19\linewidth]{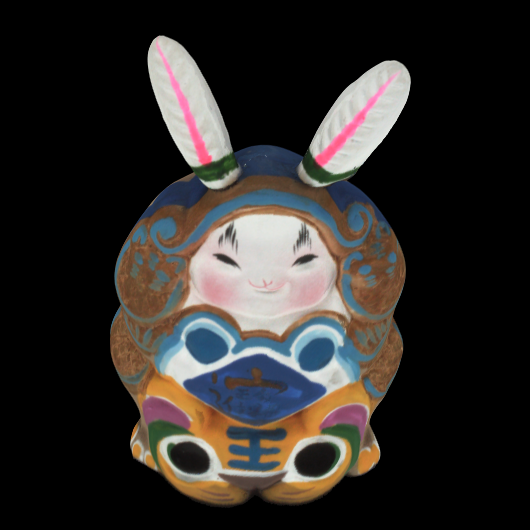}
    \includegraphics[width = 0.19\linewidth]{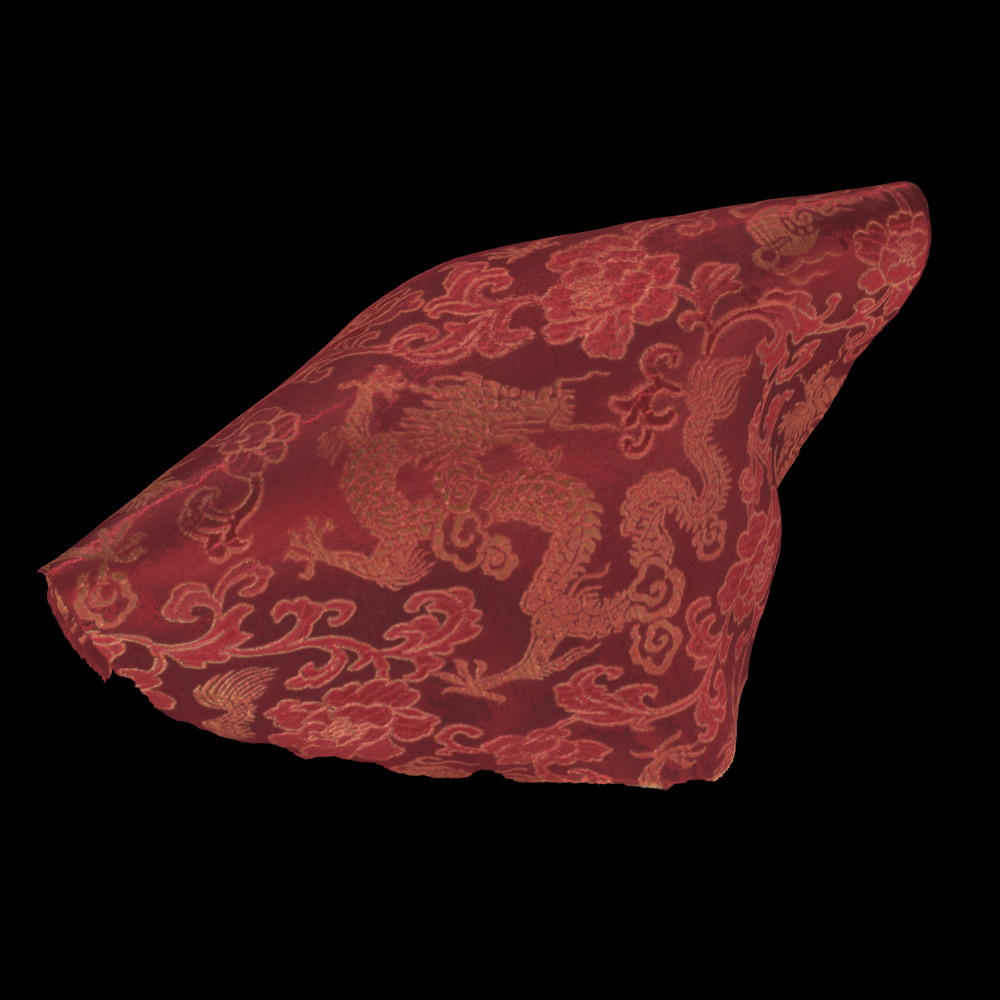}
    \includegraphics[width = 0.19\linewidth]{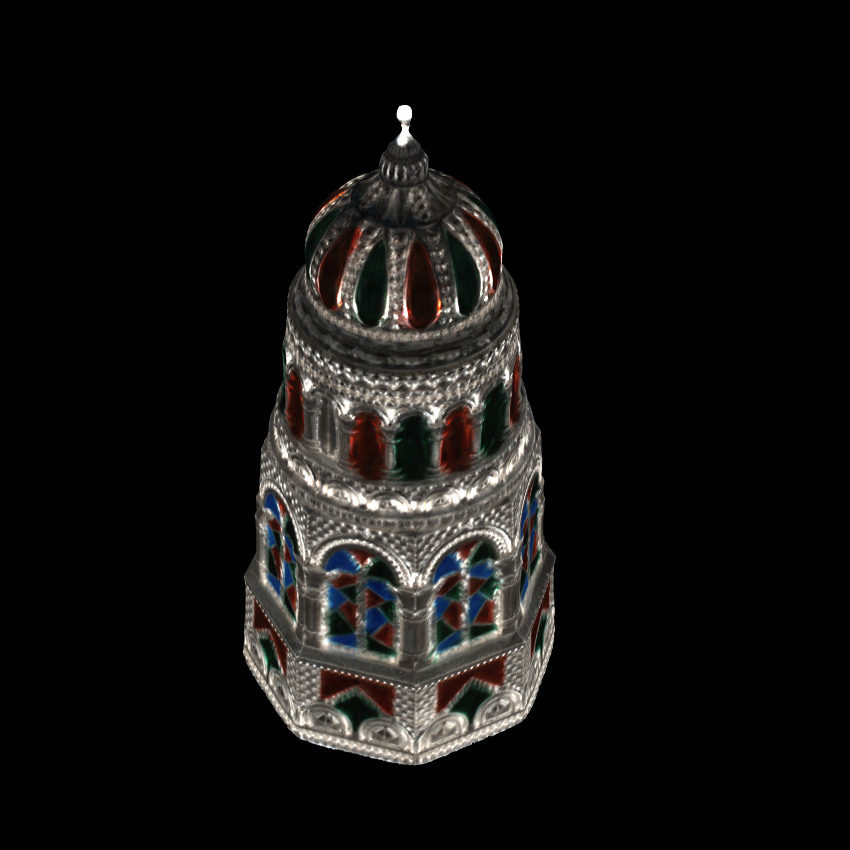}
    \includegraphics[width = 0.19\linewidth]{fig/blank.png}
  \caption{The collection of all pre-captured objects used to generate our training data. The diffuse albedos, as shown here, along with other material and geometric properties are sampled to produce attribute tensors that can be relit using arbitrary illumination conditions during network training.}
  \label{fig:data}
\end{figure}

\begin{figure}
    \includegraphics[width = 1in, angle=90]{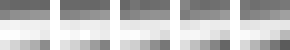}
    \includegraphics[width = 1in, angle=90]{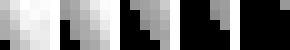}
    \includegraphics[width = 1in, angle=90]{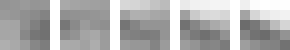}
    \includegraphics[width = 1in, angle=90]{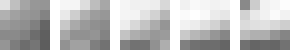}
    \includegraphics[width = 1in, angle=90]{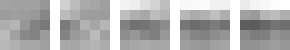}
    \includegraphics[width = 1in, angle=90]{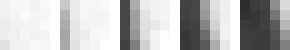}
    \includegraphics[width = 1in, angle=90]{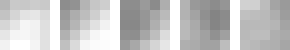}
    \includegraphics[width = 1in, angle=90]{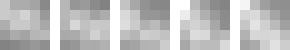}
    \includegraphics[width = 1in, angle=90]{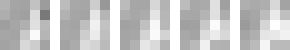}
    \includegraphics[width = 1in, angle=90]{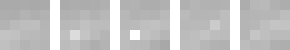}
    \includegraphics[width = 1in, angle=90]{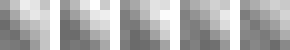}
    \includegraphics[width = 1in, angle=90]{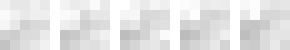}
    \includegraphics[width = 1in, angle=90]{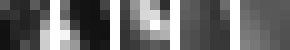}
    \includegraphics[width = 1in, angle=90]{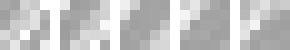}
    \includegraphics[width = 1in, angle=90]{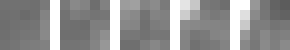}
    \includegraphics[width = 1in, angle=90]{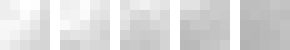}
  \caption{Examples of training tensors. They are generated by first sampling geometric and material attributes from a rotating digitized object, and then rendering with a given lighting condition. Each column of images represent a tensor of 5$\times$5$\times$5, displayed as $w_a = 5$ images of $w_s \times w_s$ = 5$\times$5 at each row.}
  \label{fig:relighting}
\end{figure}

We generate training tensors by rendering 14 pre-captured objects (\figref{fig:data}) with 3D shapes and 6D SVBRDFs represented as texture maps of GGX parameters. \figref{fig:relighting} shows some examples.

Specifically, for each object, we virtually place it on the turntable, and rotate from 0 to $360^{\circ}$ with the same angular interval as in the physical experiments. For each rotated view, we render the position/normal/tangent/BRDF parameters, and store the results as attribute maps. Next, for a valid pixel at a particular view, we assemble its \textbf{attribute tensor} by putting together its neighborhood in the spatial-angular domain from all attribute maps. Finally, this attribute tensor is used to render an input tensor of the same size for the main network under a given lighting pattern in a differentiable manner, using~\eqnref{eq:render}.

\subsection{Loss Function}
\label{sec:loss}
Ideal features in multi-view stereo should possess the following properties~\cite{zbontar2015computing,zagoruyko2015learning,tian2017l2}: (1) the features of the same point at different views are invariant; (2) the features of different points are discriminative; (3) it is efficient to compare features.

Towards these goals, we propose the following loss function defined on a batch of samples:
\begin{equation}
    L = L_{\operatorname{pos}} - \lambda L_{\operatorname{neg}},
\end{equation}
with
\begin{equation}
    L_{\operatorname{pos} / \operatorname{neg}} = \sum_{i, j \in \operatorname{pos} / \operatorname{neg}} || F(T_i) - F(T_j) ||_2.
\end{equation}
Here $T$ denotes an input tensor and $F$ the transform by our main network. $L$ consists of two terms: $L_{\operatorname{pos}}$ to encourage the view-invariance of features, and $L_{\operatorname{neg}}$ to increase spatial distinctiveness. Specifically,  $L_{\operatorname{pos}}$ measures the Euclidean feature distances between positive pairs (i.e., input tensors at two different views that belong to the same 3D surface point). $L_{\operatorname{neg}}$ measures the feature distances between negative pairs (i.e., tensors that belong to points within a spatial neighborhood of $w_{\operatorname{neg}} \times w_{\operatorname{neg}}$ at the same view). We set $\lambda$ to $0.01$ in most experiments.

To prepare a batch of $n$ samples for training, we first randomly select a surface point $p_0$ on one of the pre-captured objects, along with a pre-rendered view (\sec{sec:data}). Next, we randomly sample valid pixels in the $w_{\operatorname{neg}} \times w_{\operatorname{neg}}$ spatial neighborhood of the projection of $p_0$ at the current view, and store the corresponding surface points as $\{ p_1, p_2, ..., p_{n-1} \}$. For each $p_i$, we additionally sample another visible view along with the current one; the corresponding tensors at two views form a positive pair. On the other hand, for each pair of $\{p_i, p_j\}_{i \neq j}$, the two corresponding tensors at the current view form a negative pair.

Note that our negative pairs are "harder" than in~\cite{Kang:2021:LEARN}, as the tensors corresponding to spatially nearby points could be quite similar. This makes learning with a relative distance loss~\cite{Kang:2021:LEARN,tian2017l2} rather difficult in our pilot study. As a result, we design the current simpler loss based on absolute distances.

\subsection{Training}
Our network is implemented with PyTorch, and trained with the Adam optimizer with mini-batches of 12 and a momentum of 0.9. Xavier initialization is applied, and the learning rate of $1\times10^{-4}$. To increase the robustness in processing physical measurements, we perturb each rendered pixel in a training tensor with a multiplicative Gaussian noise ($\mu = 1, \sigma = 1\%$).

\section{Implementation Details}
At runtime, we separately transform the RGB channels of the input images, resulting in a three-channel high-dimensional feature vector at each pixel location in each view. All channels are further concatenated into a single vector, and then projected to a 3D space in the range of $[0,1]^3$ via principal component analysis. Next, the results are quantized with 8 bits and stored as conventional RGB images. For each photograph, we mask out the background using an additional image with only the back-face LEDs on~\cite{Gardner:2003:LINEAR}. Finally, the masked RGB images are fed as input to COLMAP~\cite{Schoenberger:2016:COLMAP} for 3D reconstruction.

\section{Results and Discussions}

We acquire the geometry of 4 physical objects that varies in shape and appearance. Unless otherwise noted, it takes 8 minutes to take 360 photographs of each object with a $1^{\circ}$ rotational interval, under pre-optimized lighting condition in our setup. No high-dynamic-range imaging is employed. The ground-truth shapes are obtained with a professional 3D scanner~\cite{EinScan}, after applying powder to the surfaces of the physical samples, which is necessary to reduce the adversarial specular reflections. Each captured image is downsampled to reduce noise and cropped to exclude the background to a size of about 300K effective pixels.

All computation is performed on a workstation with dual Intel
Xeon 4210 CPUs, 256GB DDR4 memory and 4 NVIDIA GeForce
RTX 3090 GPUs. It takes 71 hours to generate positive/negative pairs and compute the corresponding attribute tensors, sampled from the collection of all pre-captured objects, in a pre-processing pass (\sec{sec:data}). This results in 1.2TB of training data. Next, the network training takes 38 hours for 500K iterations. At runtime, it takes 
15 minutes to compute our DiFT features from input photographs and project to RGB images with unoptimized code, and 30 minutes to reconstruct the final 3D shape with COLMAP, a state-of-the-art non-learning-based multi-view stereo technique~\cite{Schoenberger:2016:COLMAP}.

\begin{figure}
    \includegraphics[width = 1in, angle=90]{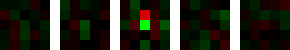}
    \includegraphics[width = 1in, angle=90]{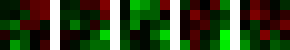}
    \includegraphics[width = 1in, angle=90]{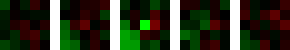}
    \includegraphics[width = 1in, angle=90]{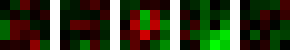}
    \includegraphics[width = 1in, angle=90]{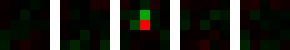}
    \includegraphics[width = 1in, angle=90]{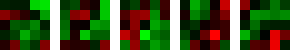}
    \includegraphics[width = 1in, angle=90]{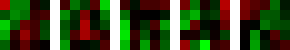}
    \includegraphics[width = 1in, angle=90]{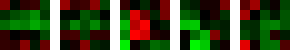}
    \includegraphics[width = 1in, angle=90]{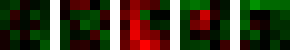}
    \includegraphics[width = 1in, angle=90]{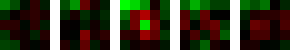}
    \includegraphics[width = 1in, angle=90]{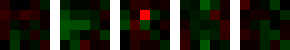}
    \includegraphics[width = 1in, angle=90]{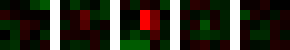}
    \includegraphics[width = 1in, angle=90]{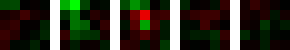}
    \includegraphics[width = 1in, angle=90]{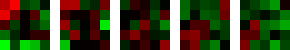}
    \includegraphics[width = 1in, angle=90]{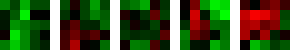}
    \includegraphics[width = 1in, angle=90]{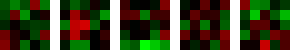}
  \caption{Visualization of our learned filters in the spatial-angular domain. Each column of images represent a filter of 5$\times$5$\times$5, displayed as $w_a = 5$ images of $w_s \times w_s$ = 5$\times$5 at each row. Positive/negative values are indicated as red/green, respectively. Only 16 filters are shown due to limited space.}
  \label{fig:filter}
\end{figure}

We visualize the weights in the first layer of the network in~\figref{fig:filter}, which are essentially learned tensor filters. It is interesting to see the variety of the filters, including edge and ring-shaped ones, as a result of DiFT training. Moreover, the non-zero weights in views other than the center one demonstrates the effectiveness of the network in exploiting differential cues in the spatial-angular domain.


\subsection{Comparisons}
\begin{figure*}
  \begin{minipage}{\textwidth}
      \begin{minipage}{0.03in}
  \hspace{0.03in}        	
      \end{minipage}	
      \begin{minipage}{\textwidth}
        \centering
      \begin{minipage}{\textwidth}
        \centering
        \begin{minipage}{0.95in}
            \centering
            {\small Features(Ours)}
        \end{minipage}		
        \begin{minipage}{0.95in}
            \centering
            {\small Features(EPFT)}
        \end{minipage}		
        \begin{minipage}{0.95in}
            \centering
            {\small Photo(MVS)}
        \end{minipage}		
        \begin{minipage}{0.95in}
            \centering
            {\small Ours}
        \end{minipage}		
        \begin{minipage}{0.95in}
            \centering
            {\small EPFT}
        \end{minipage}		
        \begin{minipage}{0.95in}
            \centering
            {\small CasMVSNet}
        \end{minipage}		
        \begin{minipage}{0.95in}
            \centering
            {\small COLMAP}
        \end{minipage}		
      \end{minipage}
  \end{minipage}       
  \end{minipage}       

\begin{minipage}{\textwidth}
    \begin{minipage}{0.03in}	
        \centering
        \rotatebox{90}{\small \textsc{Bust}}
    \end{minipage}	
    \begin{minipage}{\textwidth}
        \centering
        \begin{minipage}{0.95in}
            \includegraphics[width=0.95in]{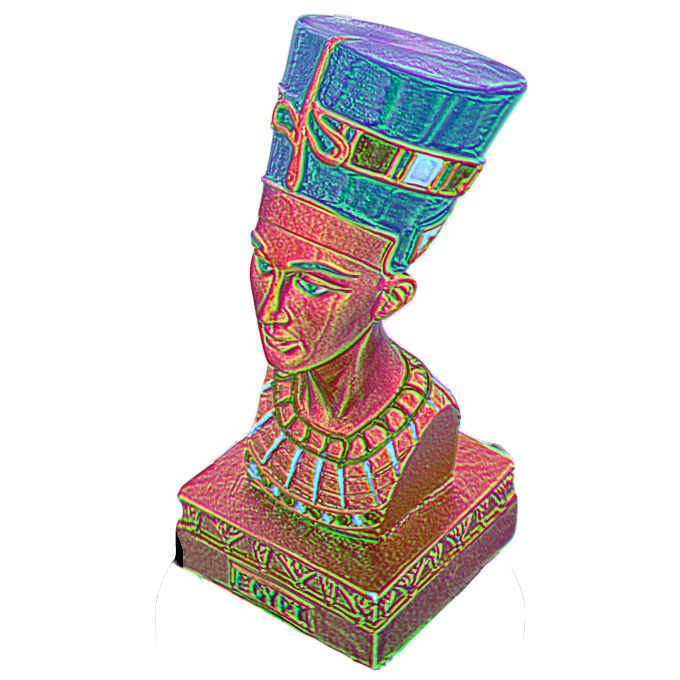}
        \end{minipage}		
        \begin{minipage}{0.95in}
            \includegraphics[width=0.95in]{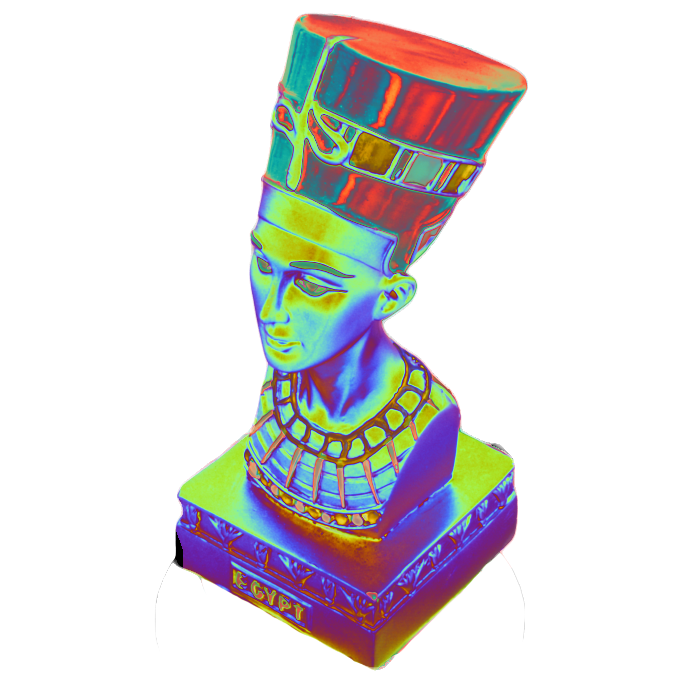}
        \end{minipage}		
        \begin{minipage}{0.95in}
            \includegraphics[width=0.95in]{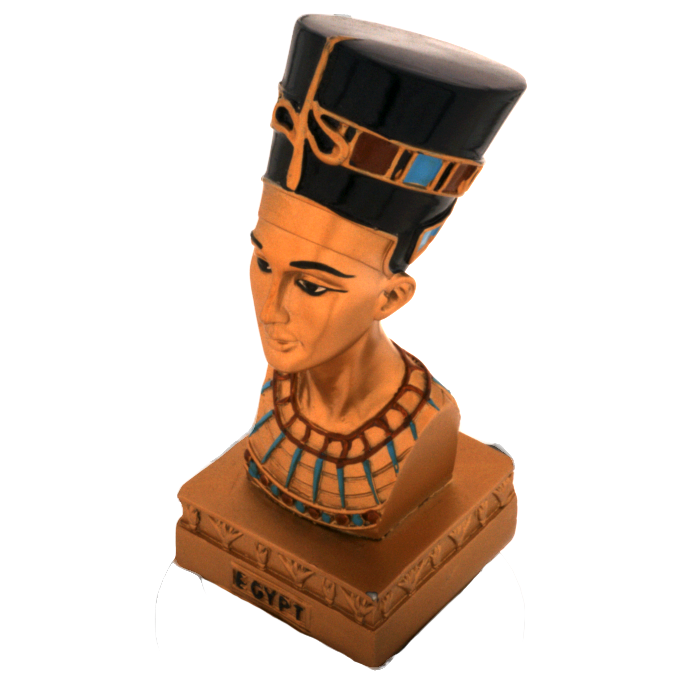}
        \end{minipage}	
        \begin{minipage}{0.95in}
            \includegraphics[width=0.95in]{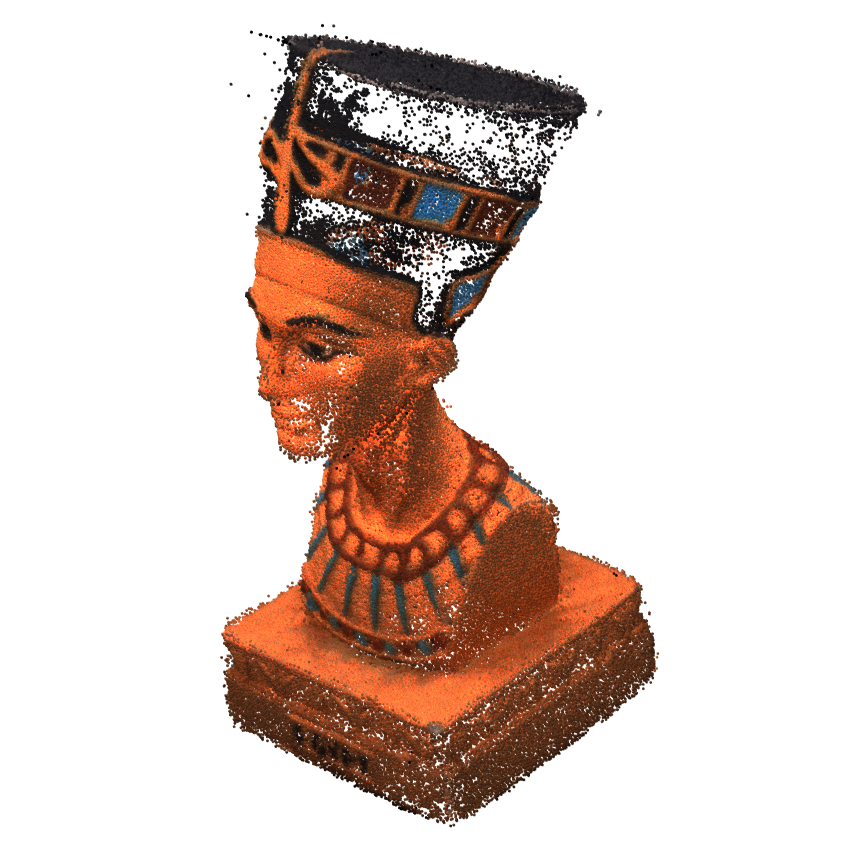}
            \put(-55,-3) {\small A:\textbf{47.8}/C:\textbf{78.4}}
        \end{minipage}		
        \begin{minipage}{0.95in}
            \includegraphics[width=0.95in]{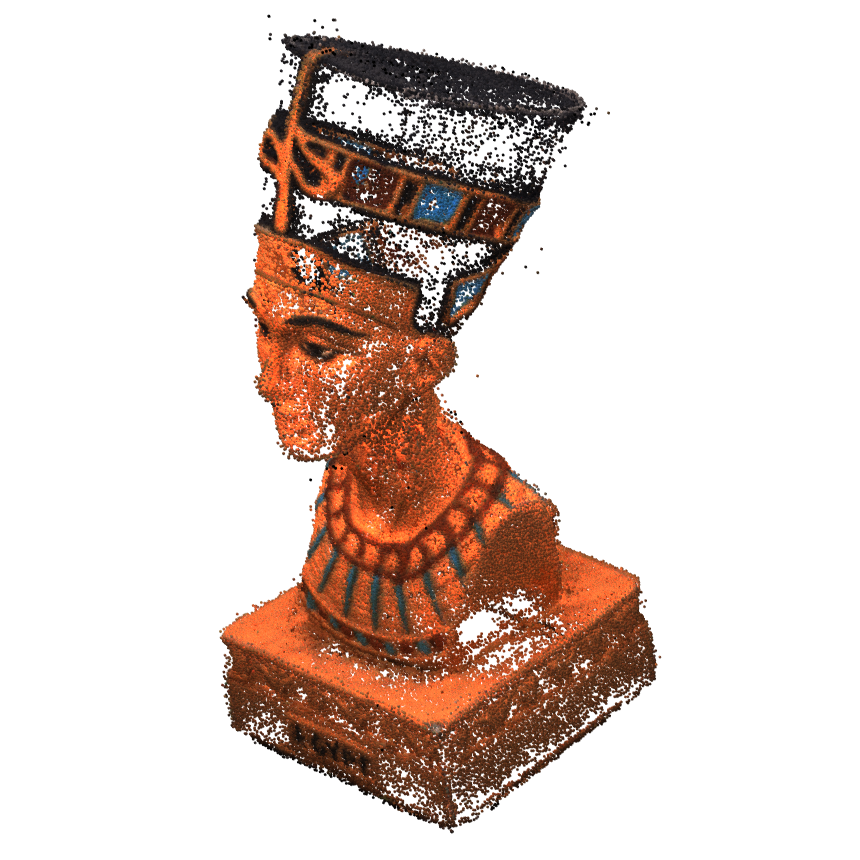}
            \put(-55,-3) {\small A:45.7/C:70.6}
        \end{minipage}		
        \begin{minipage}{0.95in}
            \includegraphics[width=0.95in]{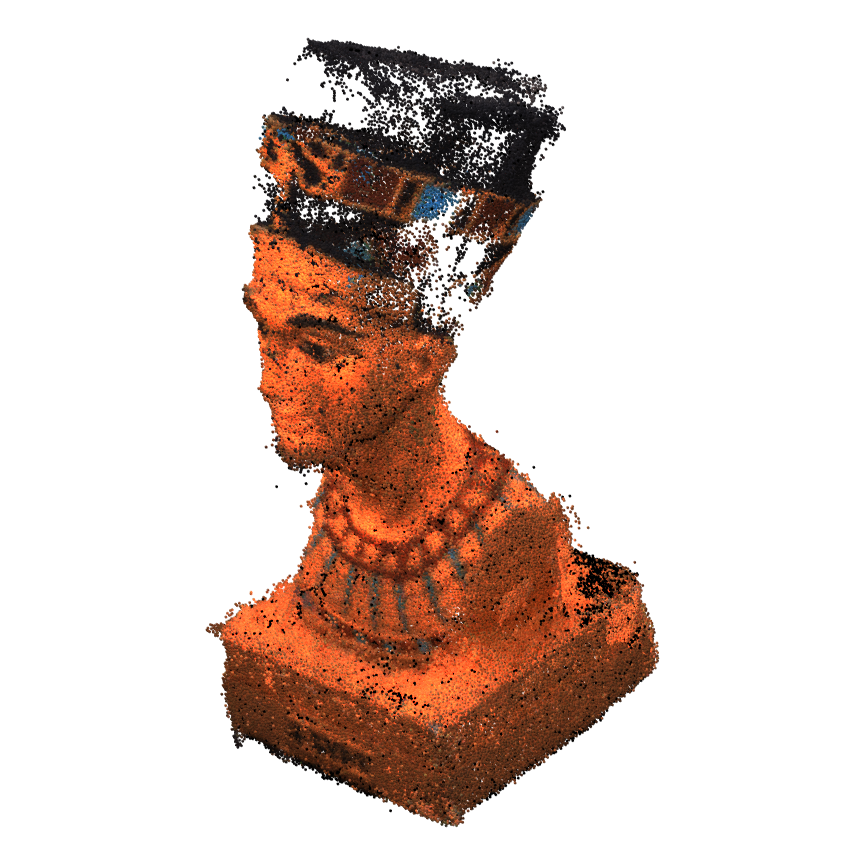}
            \put(-55,-3) {\small A:36.7/C:74.6}
        \end{minipage}		
        \begin{minipage}{0.95in}
            \includegraphics[width=0.95in]{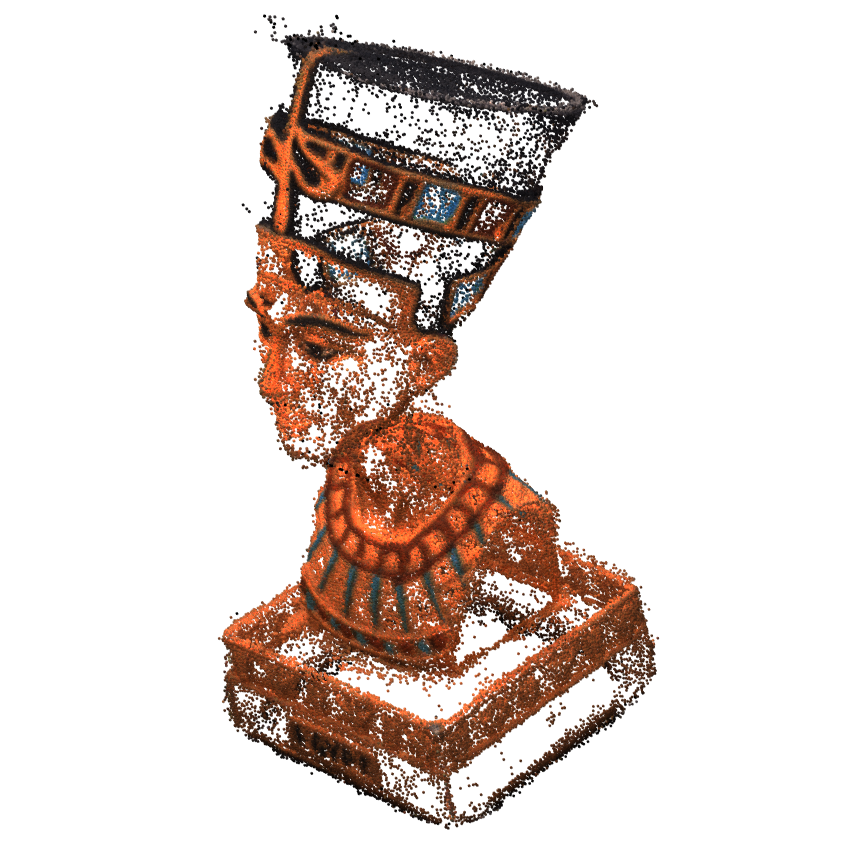}       
            \put(-55,-3) {\small A:45.9/C:55.1}
        \end{minipage}		
    \end{minipage}
\end{minipage}

\begin{minipage}{\textwidth}
    \begin{minipage}{0.03in}	
        \centering
        \rotatebox{90}{\small \textsc{Train}}
    \end{minipage}	
    \begin{minipage}{\textwidth}
        \centering
        \begin{minipage}{0.95in}
            \includegraphics[width=0.95in]{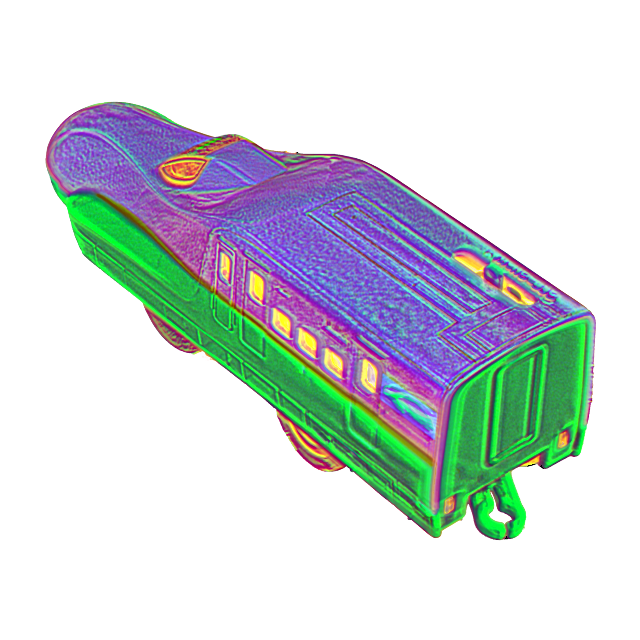}
        \end{minipage}		
        \begin{minipage}{0.95in}
            \includegraphics[width=0.95in]{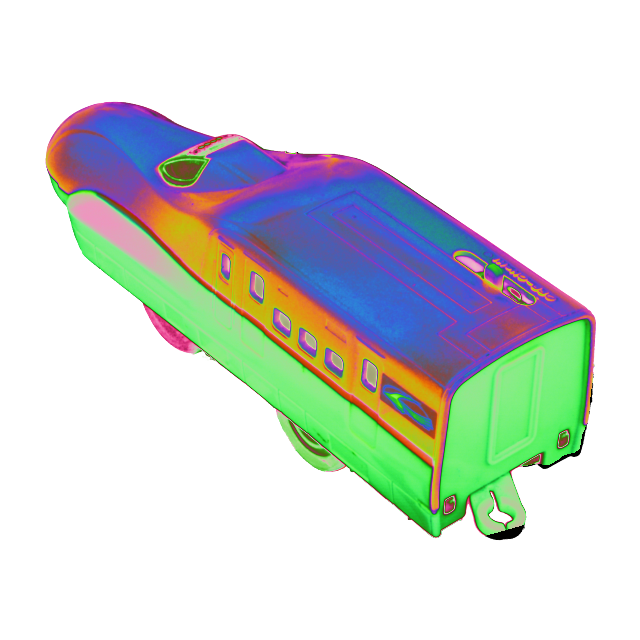}
        \end{minipage}		
        \begin{minipage}{0.95in}
            \includegraphics[width=0.95in]{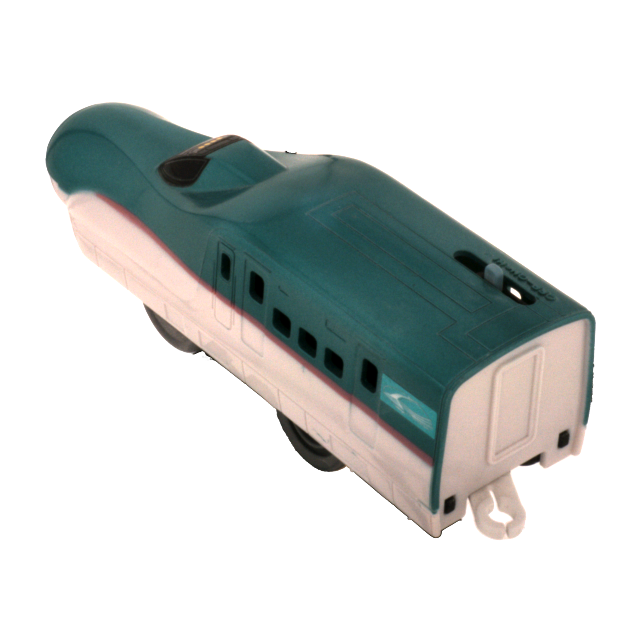}
        \end{minipage}		
        \begin{minipage}{0.95in}
            \includegraphics[width=0.95in]{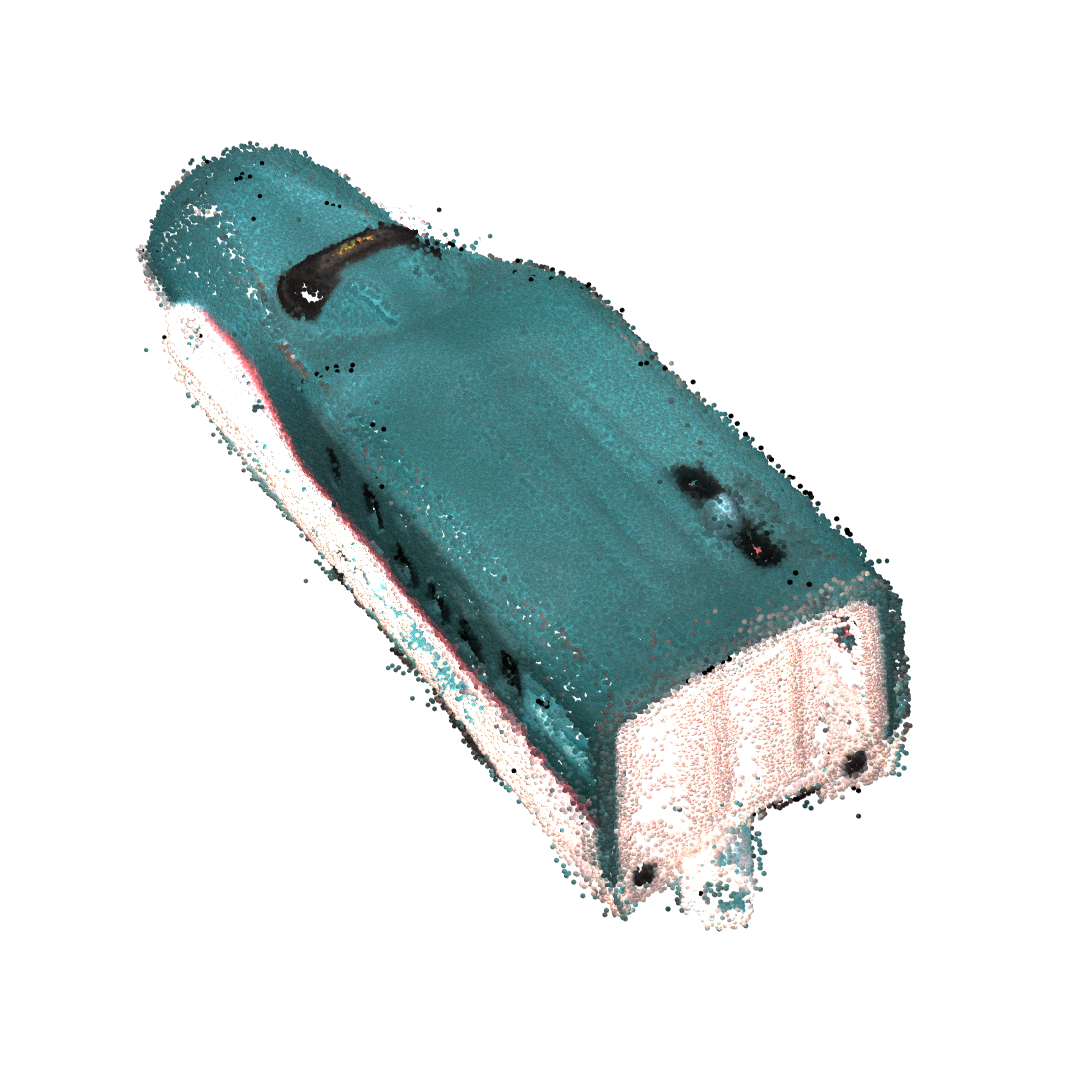}
            \put(-55,1) {\small A:\textbf{41.5}/C:49.9}
        \end{minipage}		
        \begin{minipage}{0.95in}
            \includegraphics[width=0.95in]{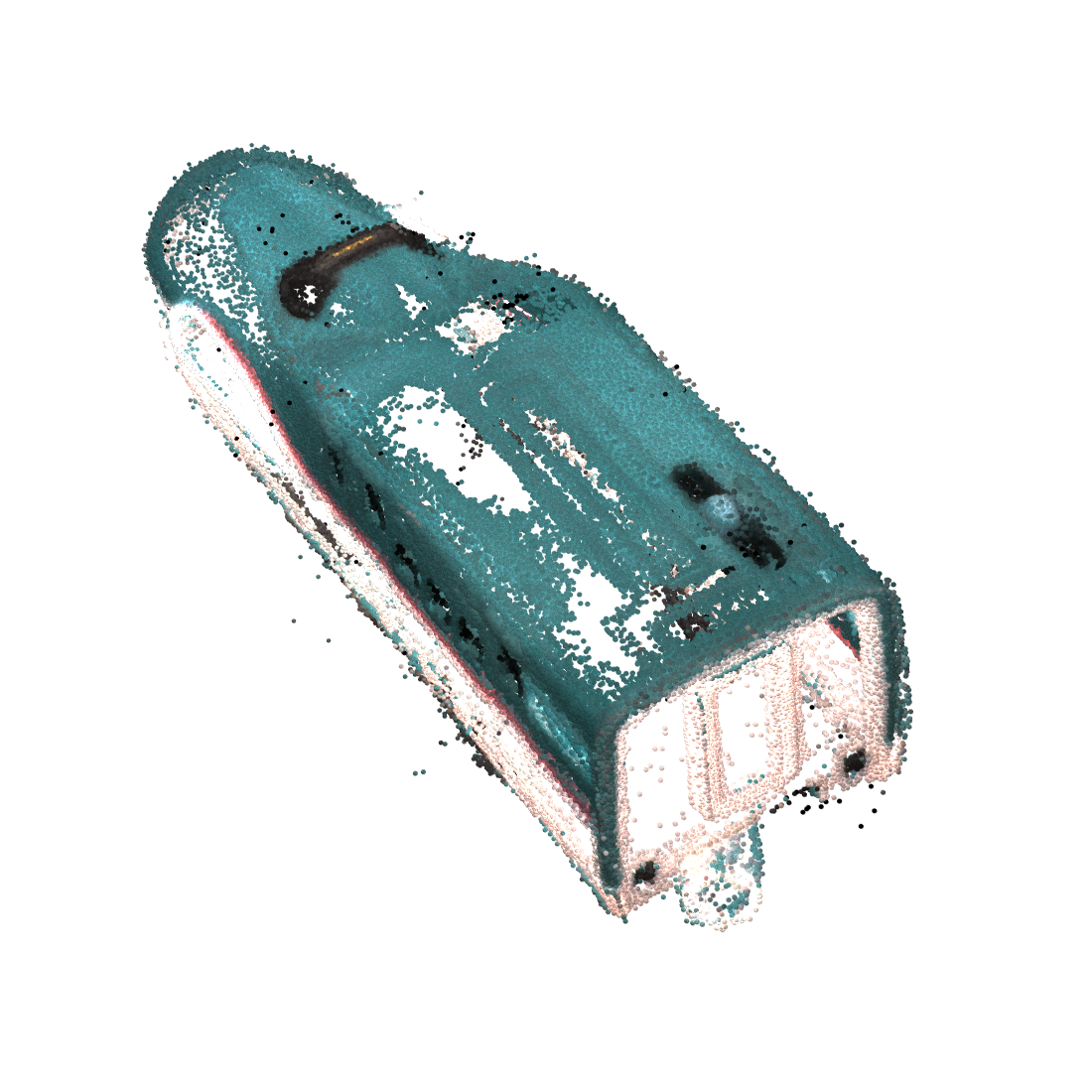}
            \put(-55,1) {\small A:40.5/C:43.7}
        \end{minipage}		
        \begin{minipage}{0.95in}
            \includegraphics[width=0.95in]{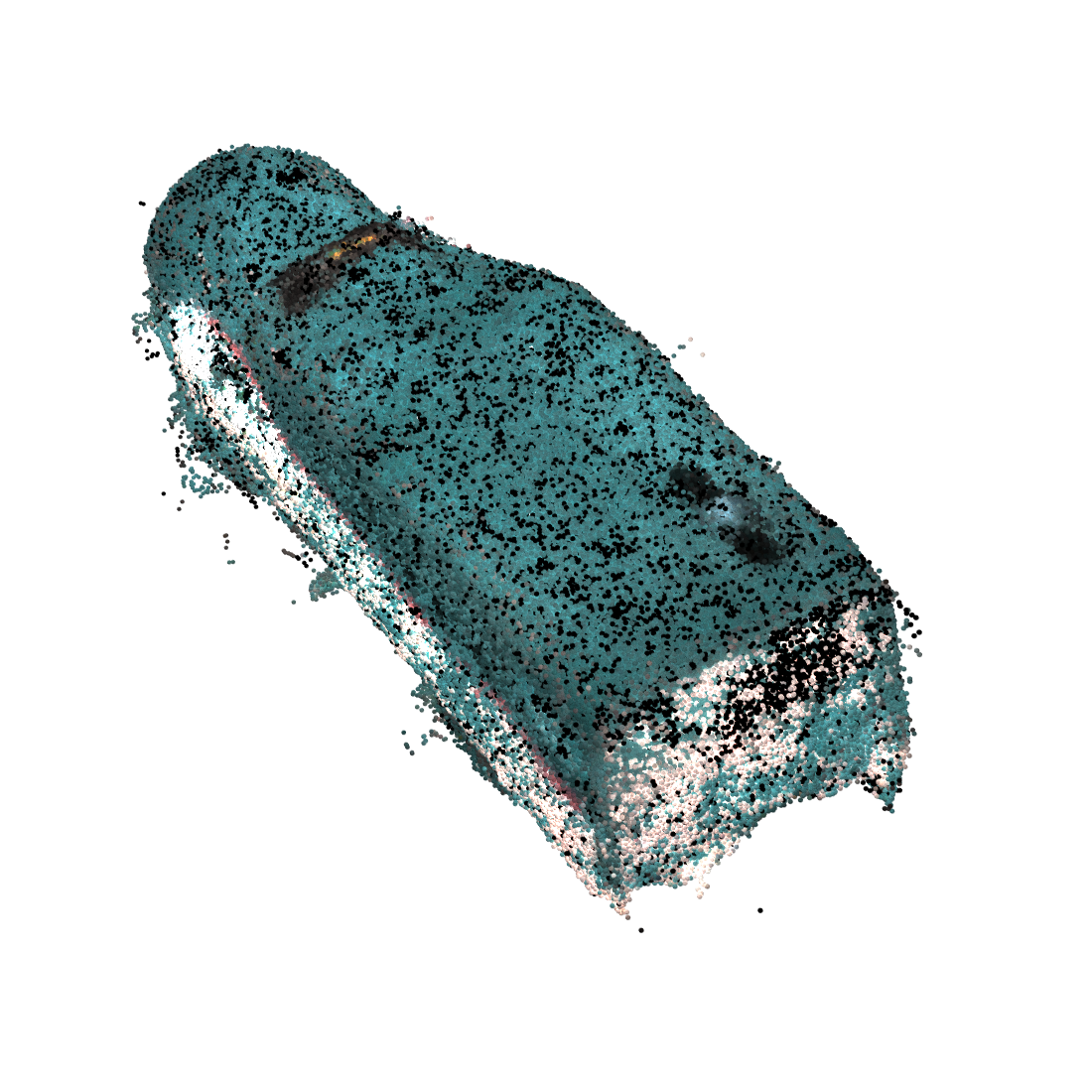}
            \put(-55,1) {\small A:34.1/C:\textbf{74.4}}
        \end{minipage}		
        \begin{minipage}{0.95in}
            \includegraphics[width=0.95in]{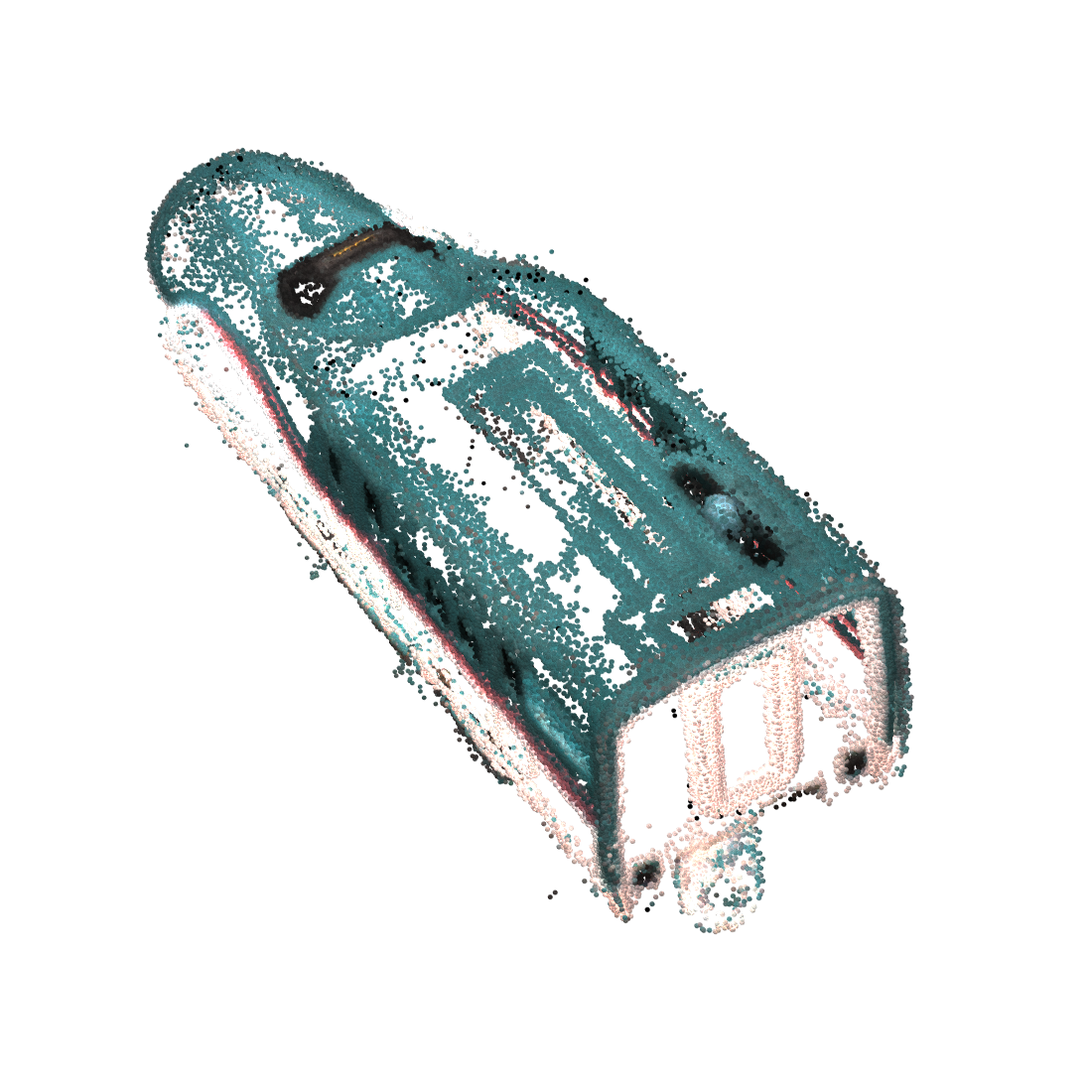}       
            \put(-55,1) {\small A:39.5/C:35.8}
        \end{minipage}	
    \end{minipage}
\end{minipage}

\begin{minipage}{\textwidth}
    \begin{minipage}{0.03in}	
        \centering
        \rotatebox{90}{\small \textsc{Cat}}
    \end{minipage}	
    \begin{minipage}{\textwidth}
        \centering
        \begin{minipage}{0.95in}
            \includegraphics[width=0.95in]{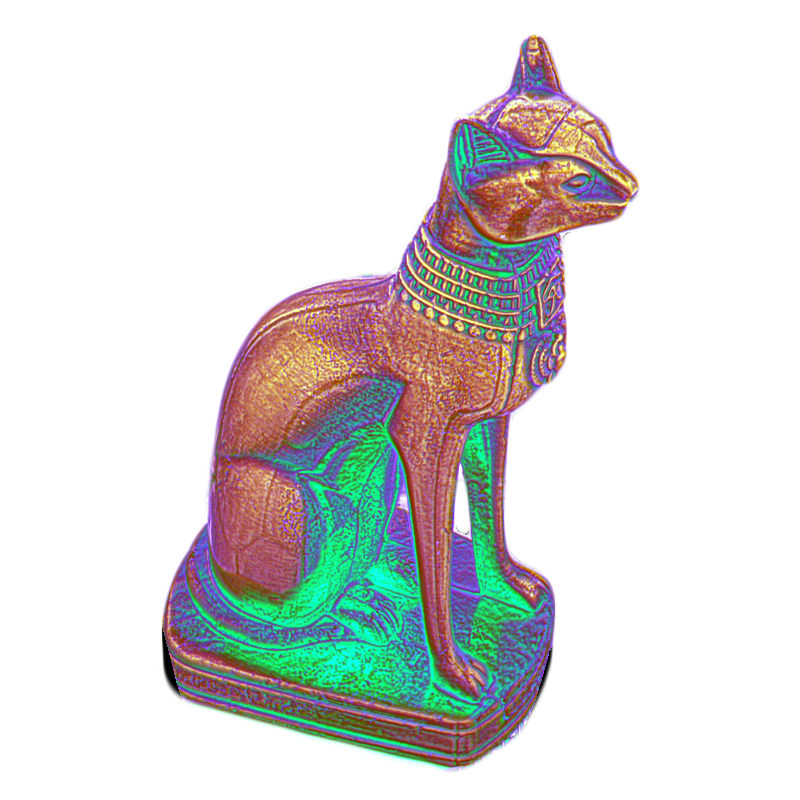}
        \end{minipage}		
        \begin{minipage}{0.95in}
            \includegraphics[width=0.95in]{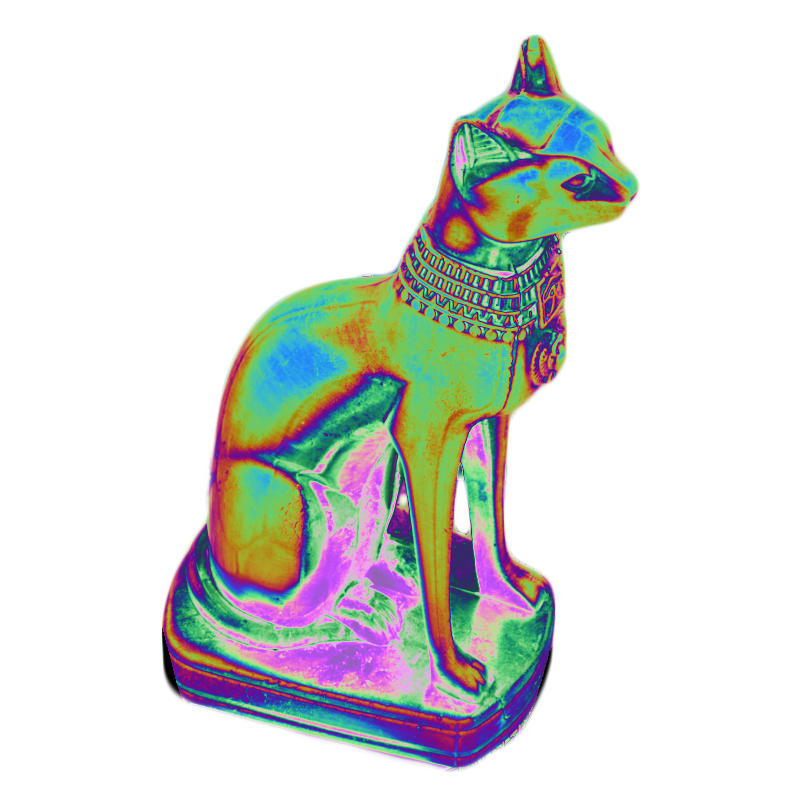}
        \end{minipage}		
        \begin{minipage}{0.95in}
            \includegraphics[width=0.95in]{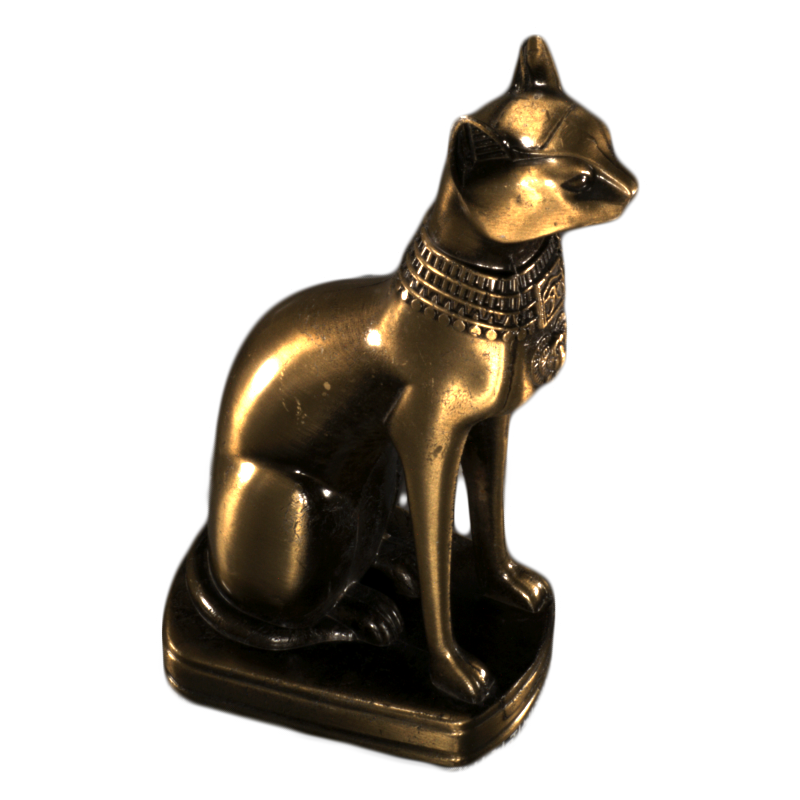}
        \end{minipage}		
        \begin{minipage}{0.95in}
            \includegraphics[width=0.95in]{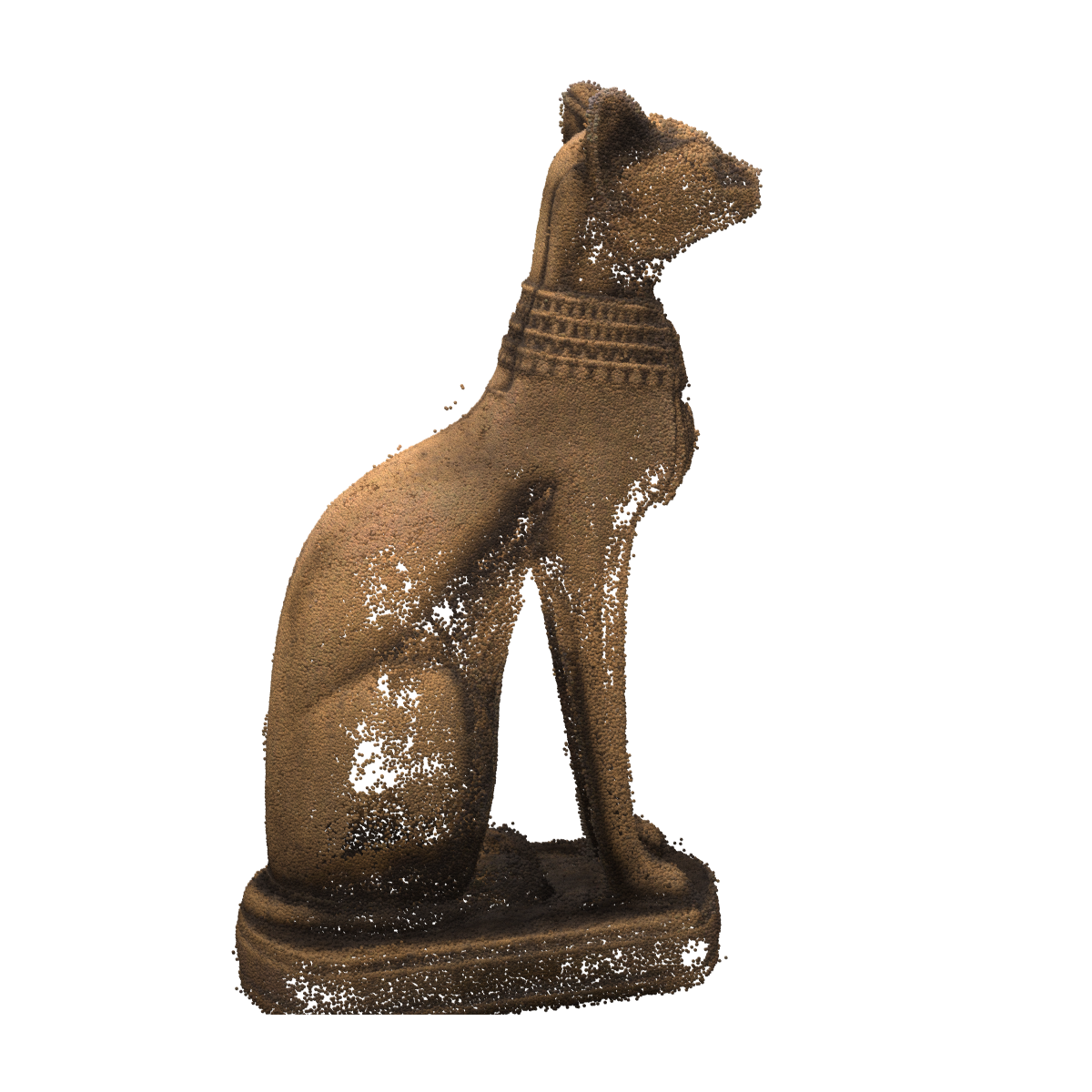}
            \put(-55,-4) {\small A:\textbf{61.6}/C:76.2}
        \end{minipage}		
        \begin{minipage}{0.95in}
            \includegraphics[width=0.95in]{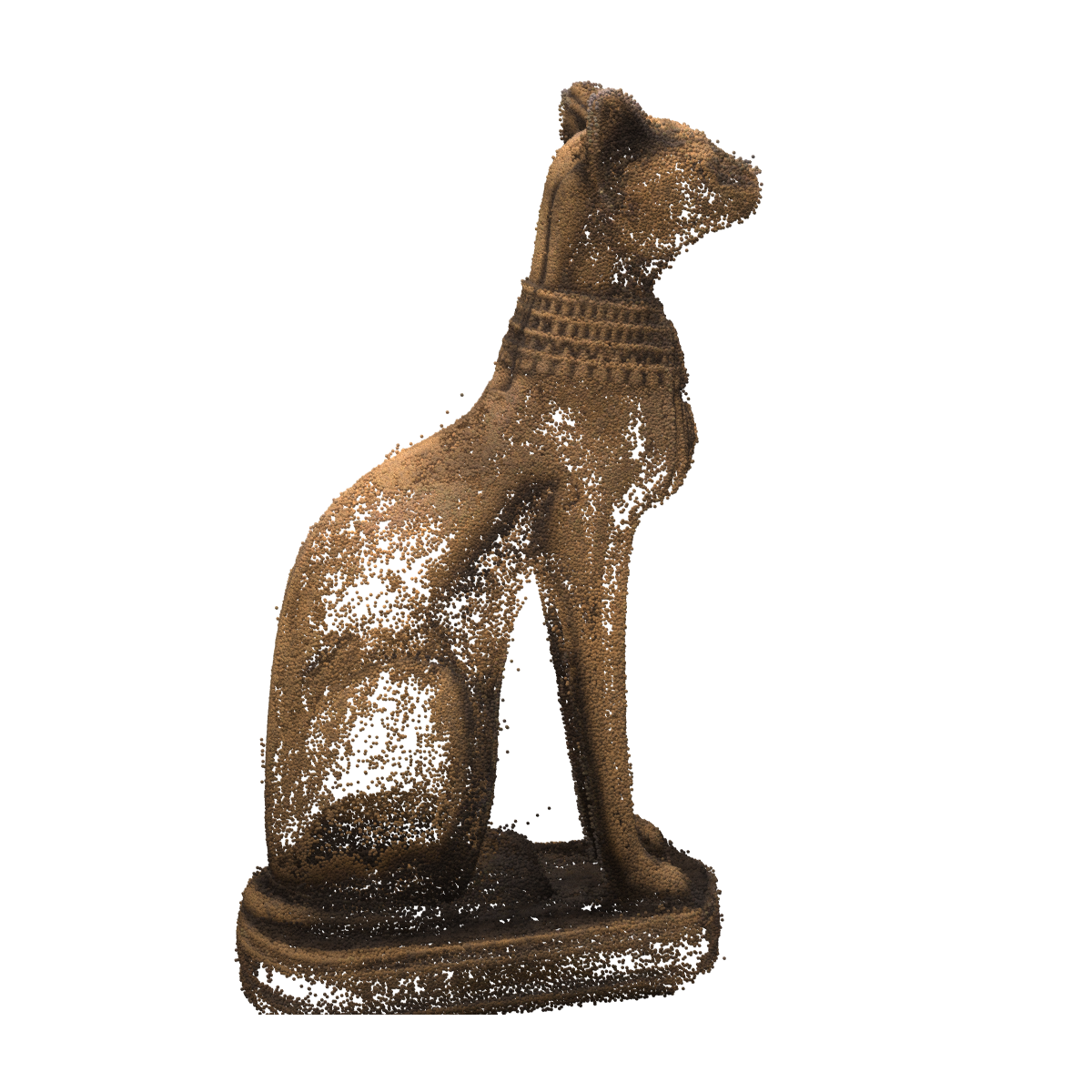}
            \put(-55,-4) {\small A:55.7/C:66.0}
        \end{minipage}		
        \begin{minipage}{0.95in}
            \includegraphics[width=0.95in]{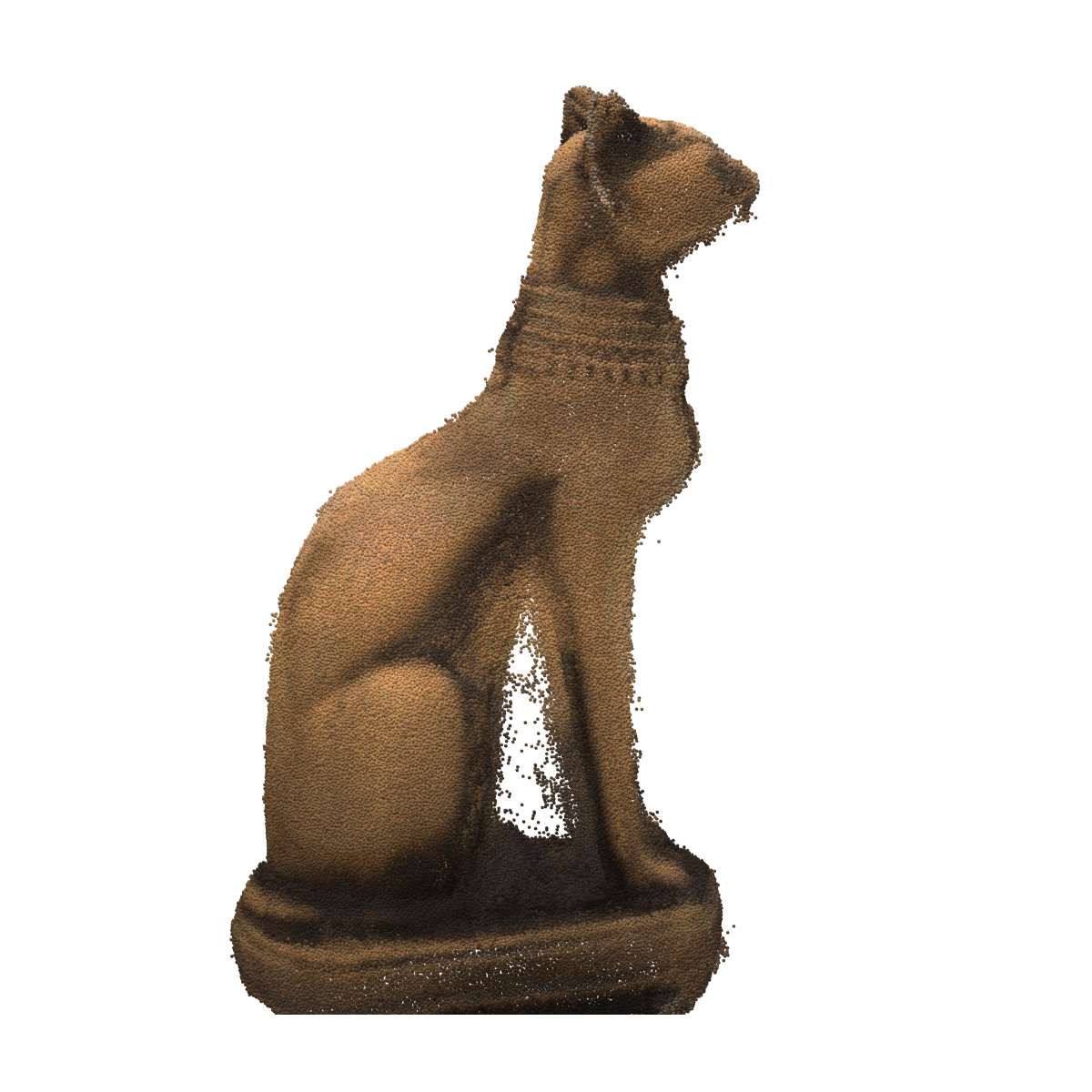}
            \put(-55,-4) {\small A:44.9/C:\textbf{86.8}}
        \end{minipage}		
        \begin{minipage}{0.95in}
            \includegraphics[width=0.95in]{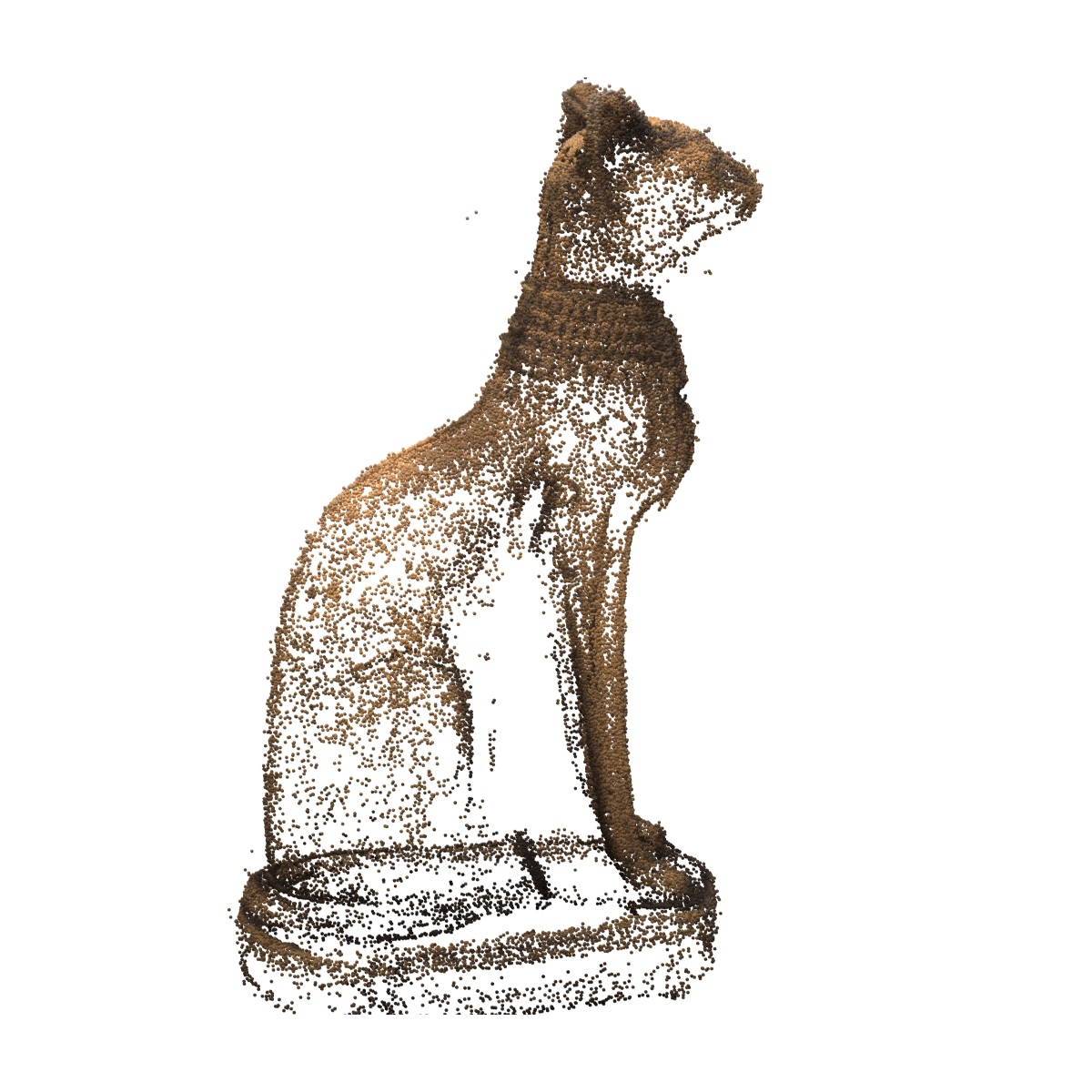}  
            \put(-55,-4) {\small A:31.7/C:24.6}
        \end{minipage}			
    \end{minipage}
\end{minipage}

\begin{minipage}{\textwidth}
    \begin{minipage}{0.03in}	
        \centering
        \rotatebox{90}{\small \textsc{Baymax}}
    \end{minipage}	
    \begin{minipage}{\textwidth}
        \centering
        \begin{minipage}{0.95in}
            \includegraphics[width=0.95in]{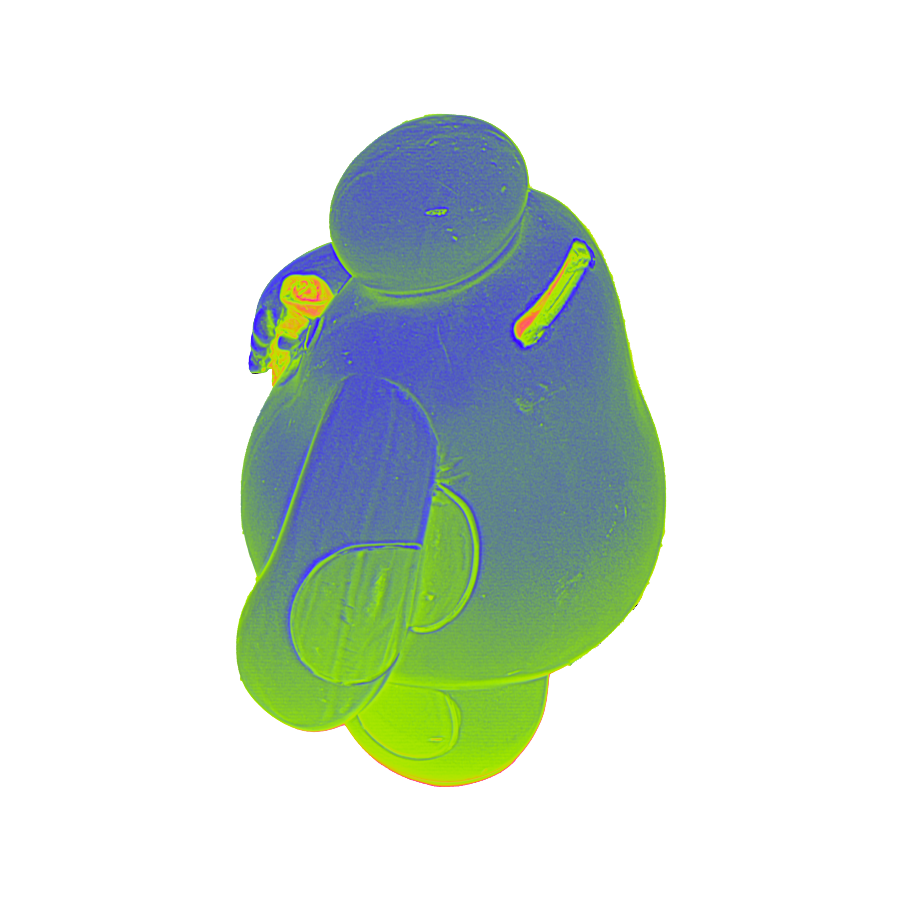}
        \end{minipage}		
        \begin{minipage}{0.95in}
            \includegraphics[width=0.95in]{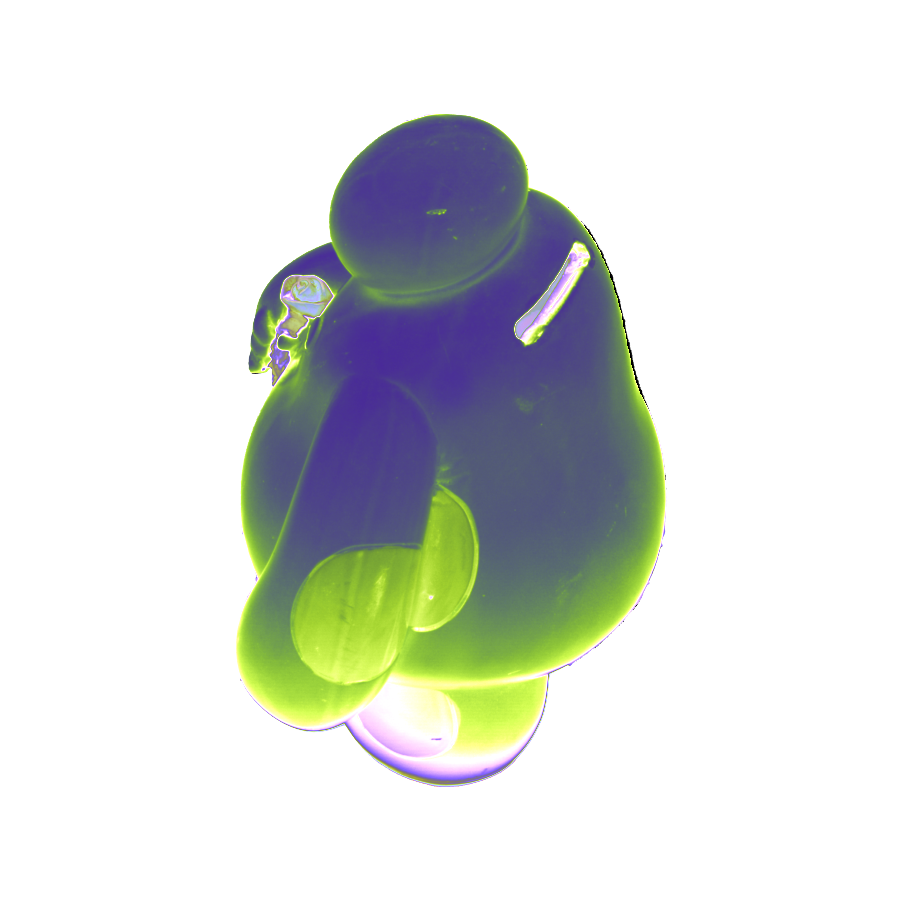}
        \end{minipage}		
        \begin{minipage}{0.95in}
            \includegraphics[width=0.95in]{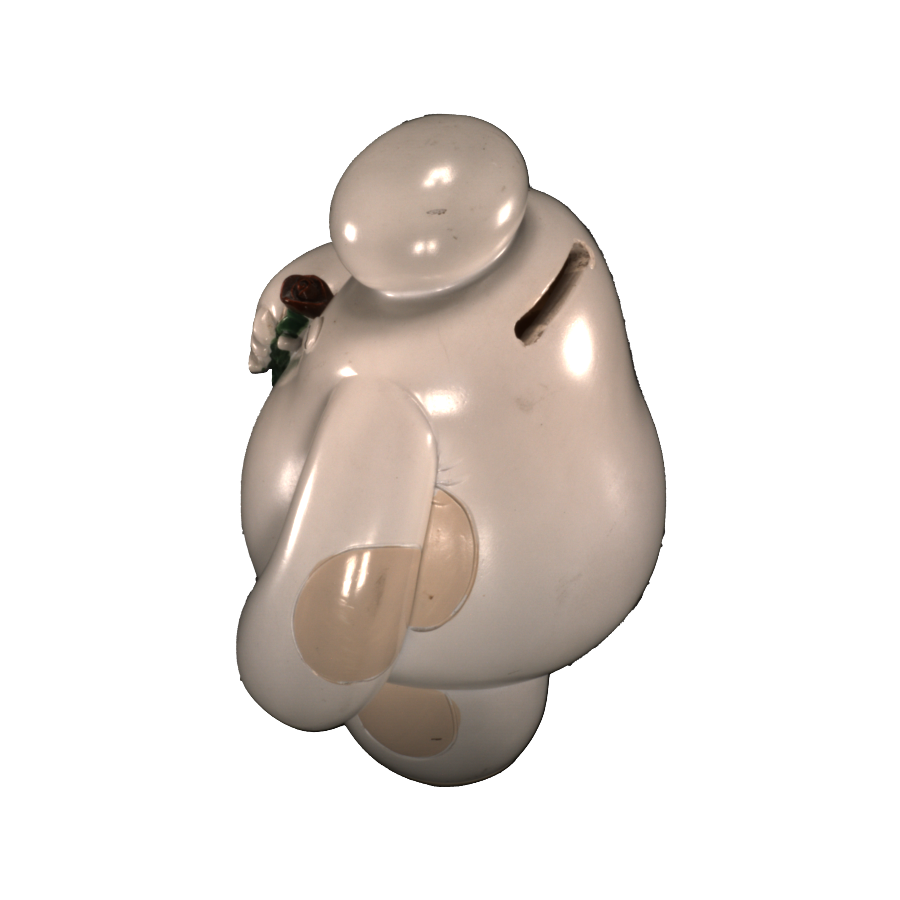}
        \end{minipage}
        \begin{minipage}{0.95in}
            \includegraphics[width=0.95in]{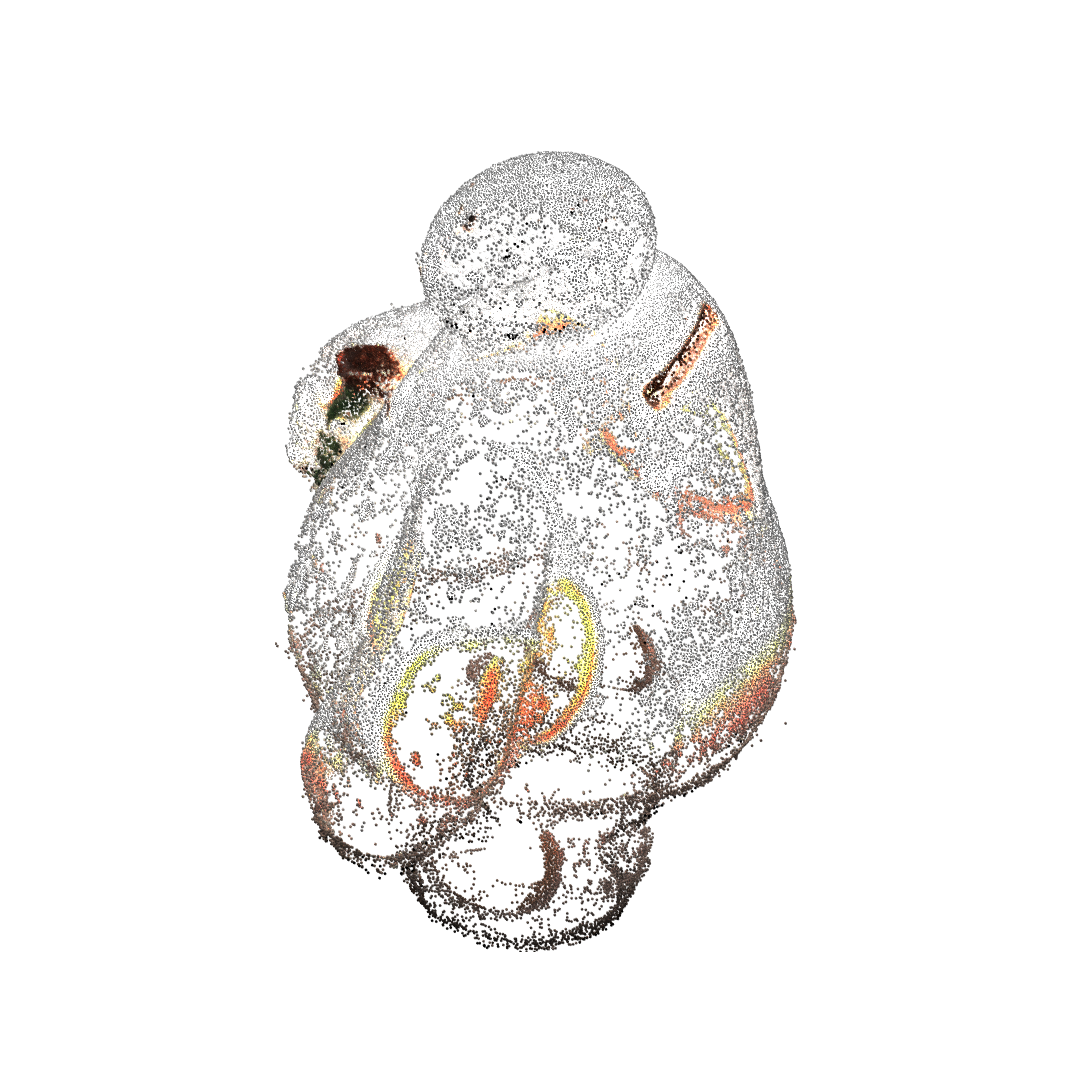}
            \put(-55,-1) {\small A:\textbf{41.9}/C:41.6}
        \end{minipage}		
        \begin{minipage}{0.95in}
            \includegraphics[width=0.95in]{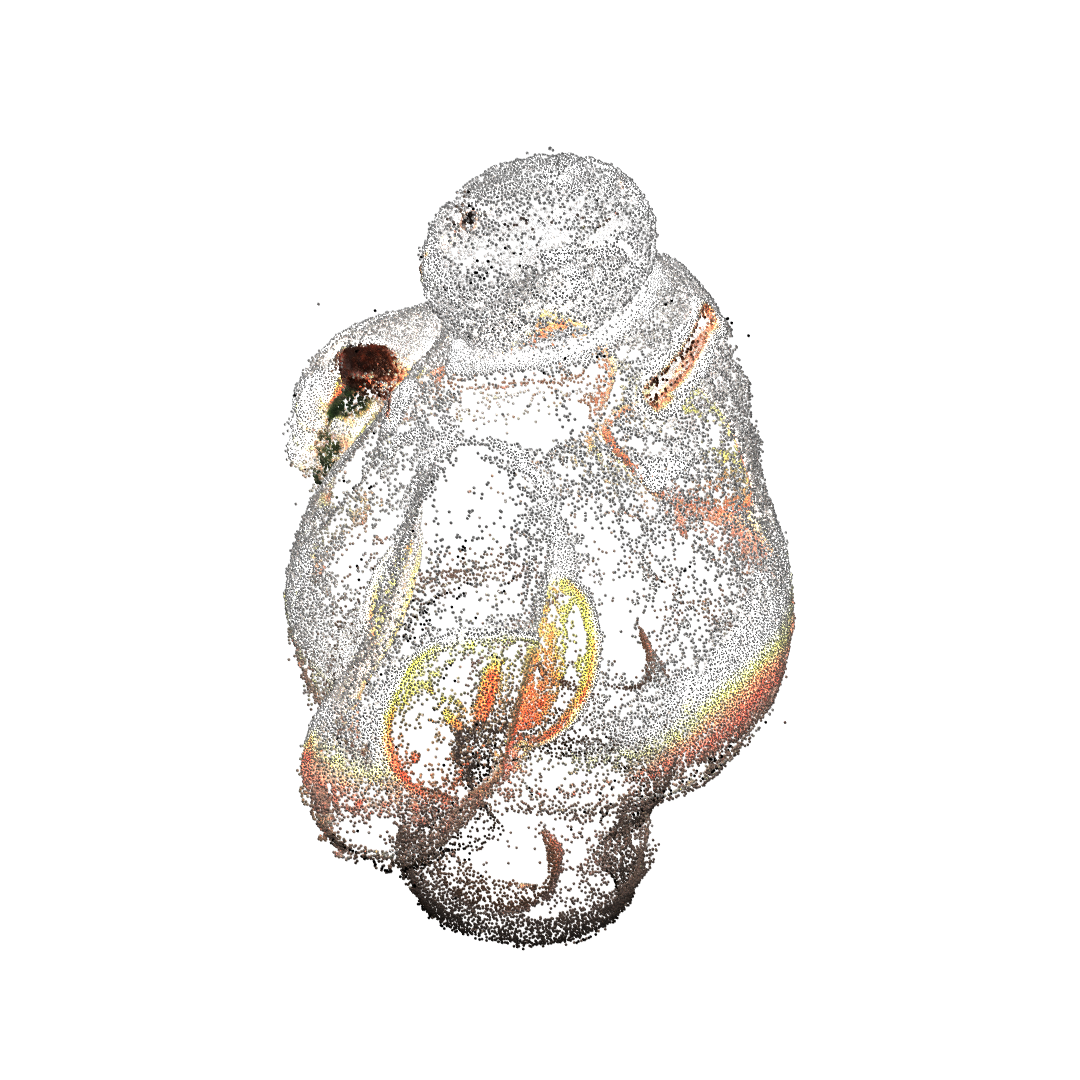}
            \put(-55,-1) {\small A:38.1/C:37.9}
        \end{minipage}		
        \begin{minipage}{0.95in}
            \includegraphics[width=0.95in]{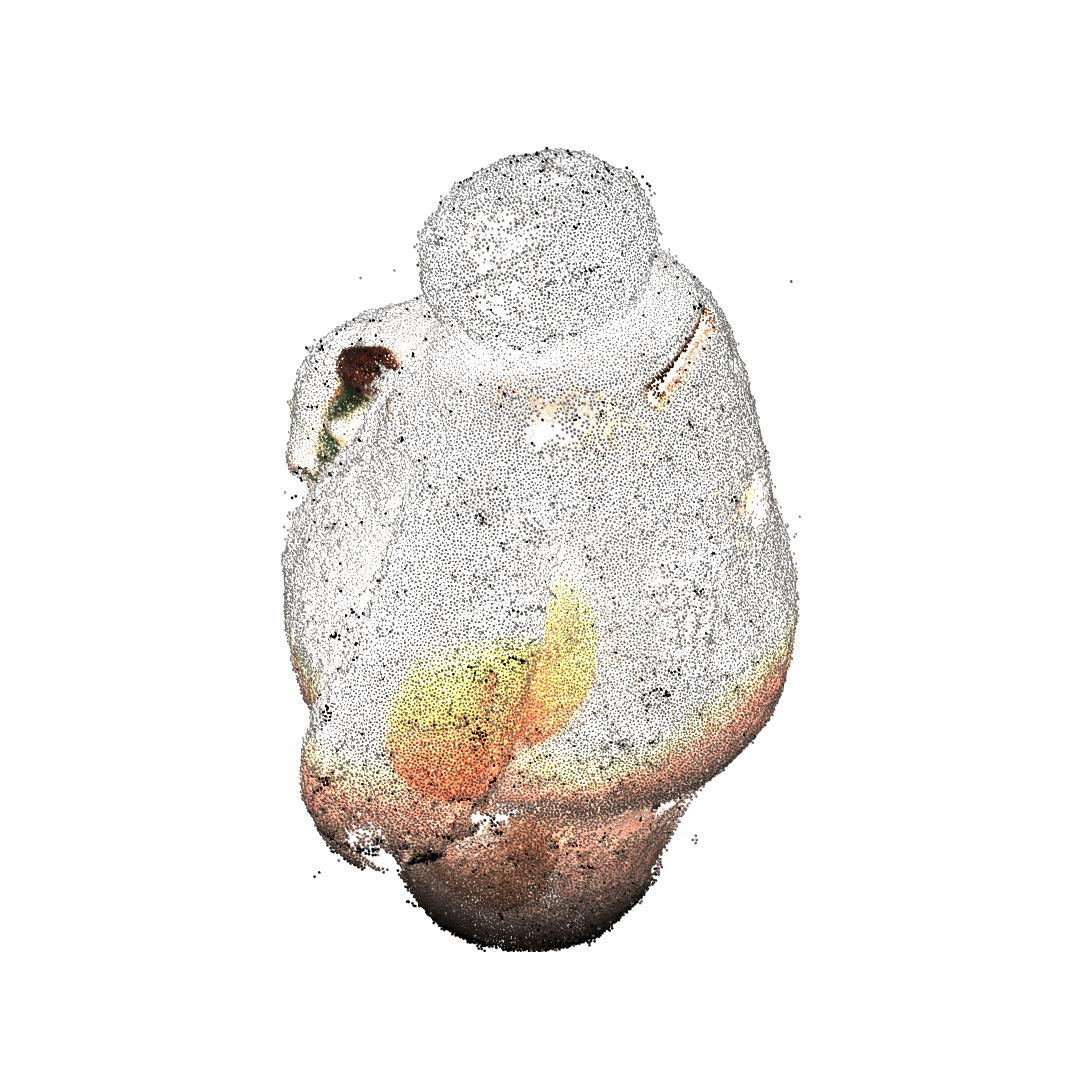}
            \put(-55,-1) {\small A:33.8/C:\textbf{73.0}}
        \end{minipage}		
        \begin{minipage}{0.95in}
            \includegraphics[width=0.95in]{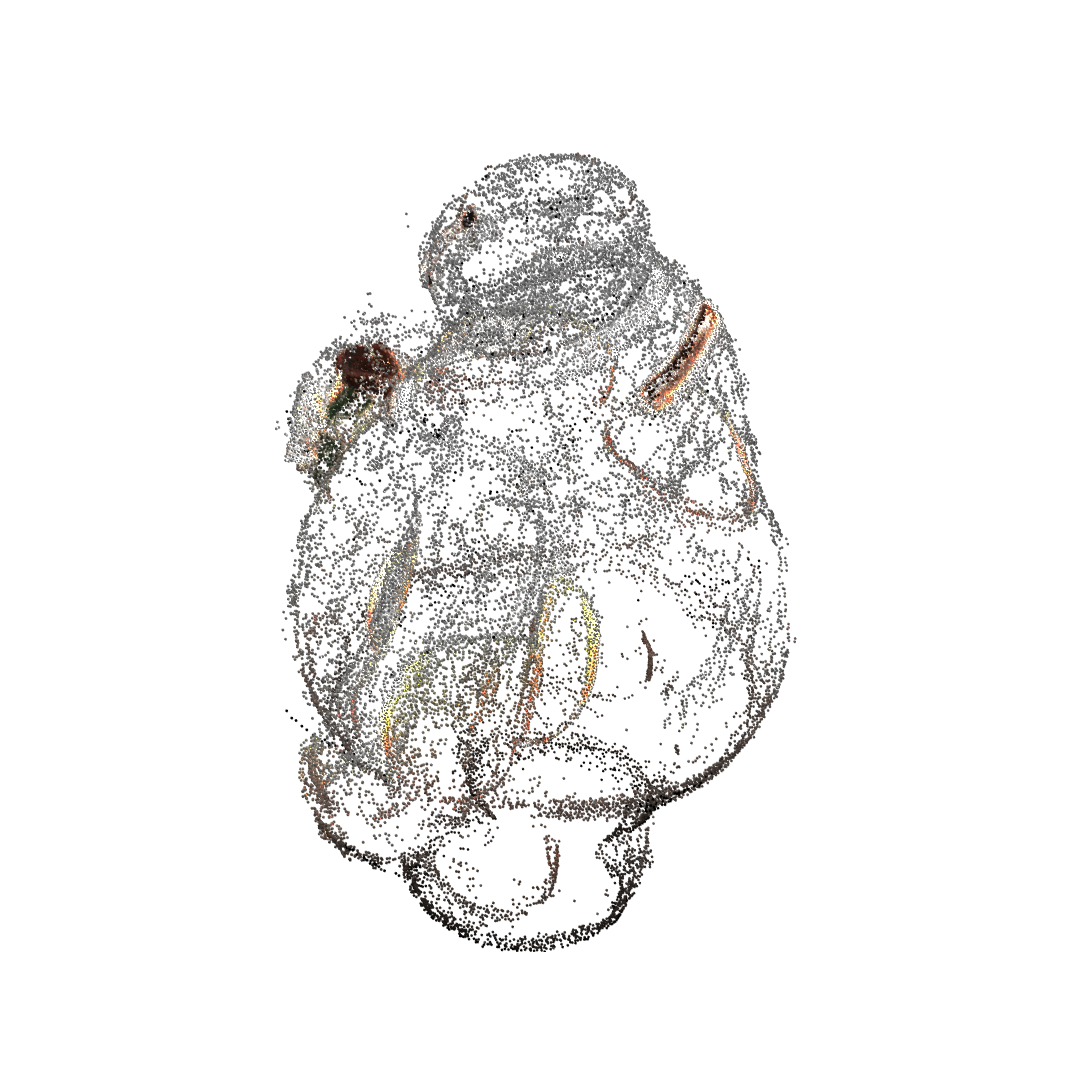}       
            \put(-55,-1) {\small A:28.2/C:16.2}
        \end{minipage}			
    \end{minipage}
\end{minipage}

  \caption{Comparisons with related work. From the left column to right, our features, EPFT features~\cite{Kang:2021:LEARN}, photographs with a fixed lighting condition, the results using DiFT, EPFT, CasMVSNet~\cite{gu2020cascade} and COLMAP~\cite{Schoenberger:2016:COLMAP}. The third column is input to both CasMVSNet and COLMAP, which uses a full-on pattern for the top two objects and a 4-point lighting for the rest.}
  \label{fig:comp}
\end{figure*}
We compare in~\figref{fig:comp} the reconstruction results of 4 physical objects captured with our device, using DiFT against three closely related methods. EPFT is the closest work to ours, which learns efficient features from photometric information~\cite{Kang:2021:LEARN}. For a fair comparison, we train EPFT with a single lighting pattern, the same as in DiFT. CasMVSNet is a state-of-the-art deep-learning-based multi-view stereo technique~\cite{gu2020cascade}. COLMAP, on the other hand, is the representative non-learning-based stereo work~\cite{Schoenberger:2016:COLMAP}, as well as the back-end of EPFT and DiFT. All four approaches take as input 360 photographs. DiFT and EPFT employ a pre-optimized lighting pattern during acquisition. For CasMVSNet and COLMAP, one of two commonly used lighting conditions is employed for different objects: a full-on pattern physically reduces the view variance of complex appearance, while a 4-point pattern introduces more shading cues. Please refer to the accompanying video for the input photographs of DiFT along with the corresponding feature maps at all captured views. Note that the same set of parameters are used in all 3D reconstruction experiments.

In the figure, quantitative errors in accuracy/completeness (\%) at a 0.5mm threshold are reported (as A/C). For accuracy, we achieve consistently the highest scores among all approaches, due to the efficient exploitation of useful differential cues in the spatial-angular domain. For completeness, we rank first in \textsc{Bust}, and are second to CasMVSNet for other objects. The reason is that in our experiments, CasMVSNet tends to output overly dense points that are visually pleasing, but with a lower accuracy. A visual comparison of all results also confirms this point. Moreover, compared with EPFT trained with relatively independent negative pairs, our negative samples are from a close spatial neighborhood, leading to desirable higher-frequency features as shown in the figure, for enhanced reconstruction.



\subsection{Evaluations}
\begin{figure}
    \begin{minipage}{\columnwidth}
        \centering
        \begin{minipage}{0.03\columnwidth}	
            \centering
            \rotatebox{90}{\small $\lambda = 0.1$}
        \end{minipage}	
        \begin{minipage}{0.92\columnwidth}	
            \centering
            \includegraphics[width = 0.27\linewidth]{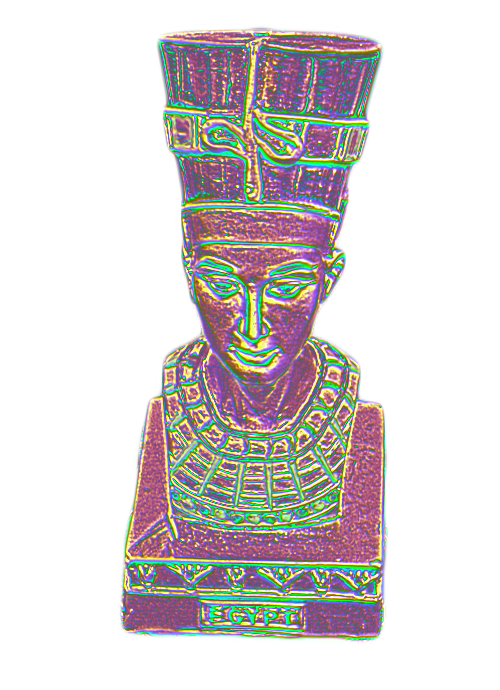}
            \includegraphics[width = 0.27\linewidth]{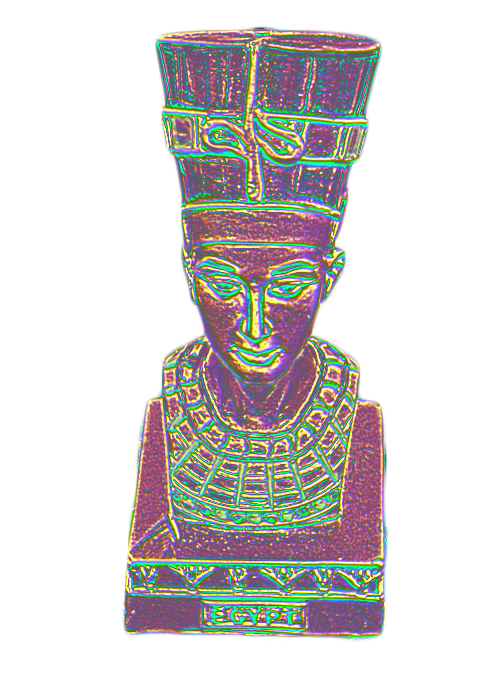}
            \includegraphics[width = 0.27\linewidth]{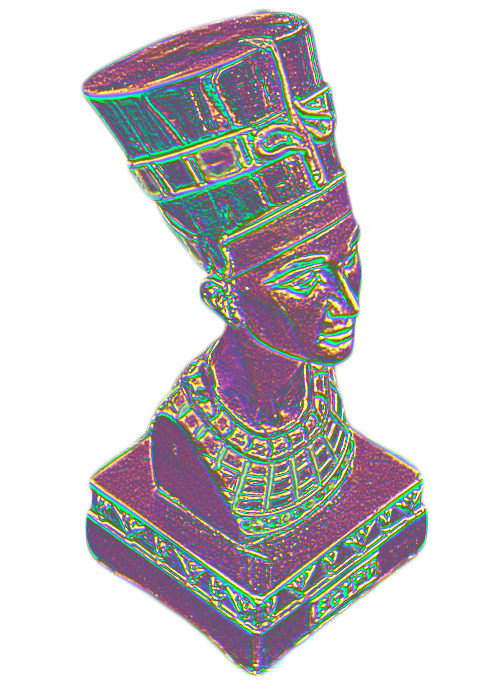}
        \end{minipage}	
        \begin{minipage}{0.03\columnwidth}	
            \centering
            \rotatebox{90}{\small A=43.0/C=73.1}
        \end{minipage}	
    \end{minipage}	

    \begin{minipage}{\columnwidth}
        \centering
        \begin{minipage}{0.03\columnwidth}	
            \centering
            \rotatebox{90}{\small $\lambda = 0.01$}
        \end{minipage}	
        \begin{minipage}{0.92\columnwidth}	
            \centering
            \includegraphics[width = 0.27\linewidth]{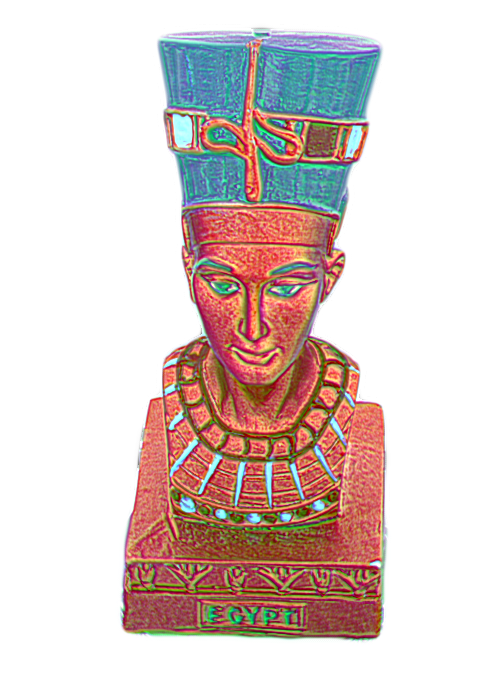}
            \includegraphics[width = 0.27\linewidth]{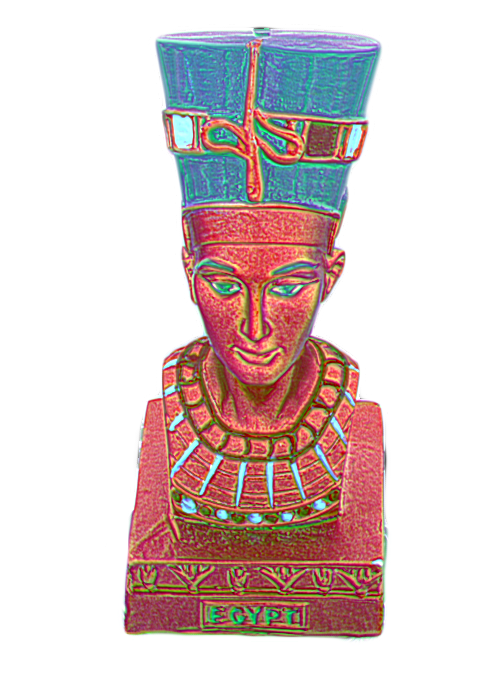}
            \includegraphics[width = 0.27\linewidth]{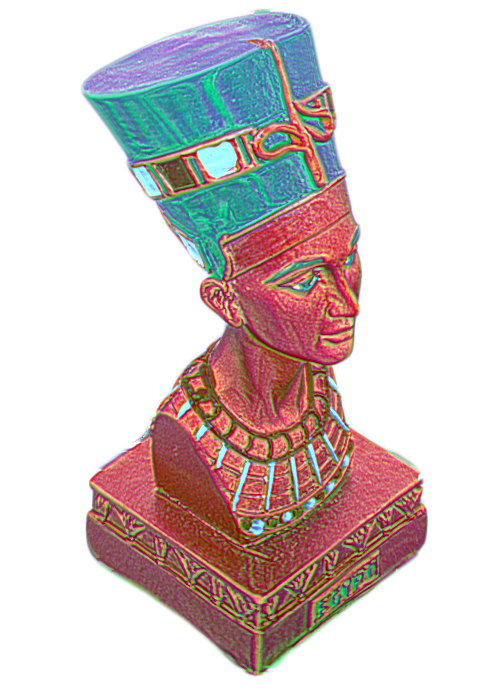}
        \end{minipage}	
        \begin{minipage}{0.03\columnwidth}	
            \centering
            \rotatebox{90}{\small A=47.8/C=78.4}
        \end{minipage}	
    \end{minipage}	

    \begin{minipage}{\columnwidth}
        \centering
        \begin{minipage}{0.03\columnwidth}	
            \centering
            \rotatebox{90}{\small $\lambda = 0.001$}
        \end{minipage}	
        \begin{minipage}{0.92\columnwidth}	
            \centering
            \includegraphics[width = 0.27\linewidth]{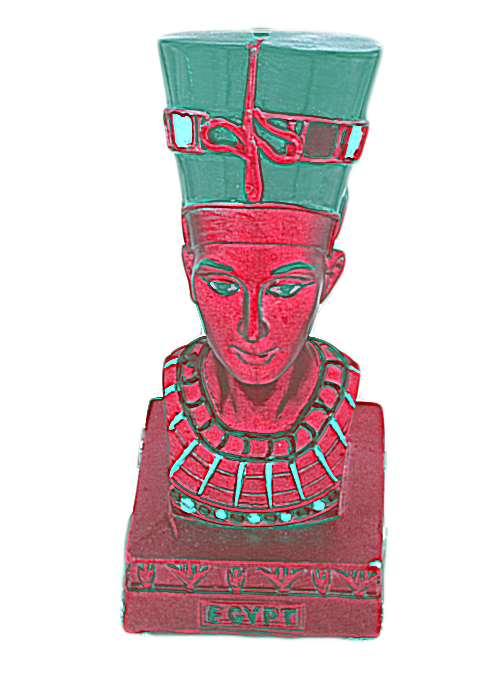}
            \includegraphics[width = 0.27\linewidth]{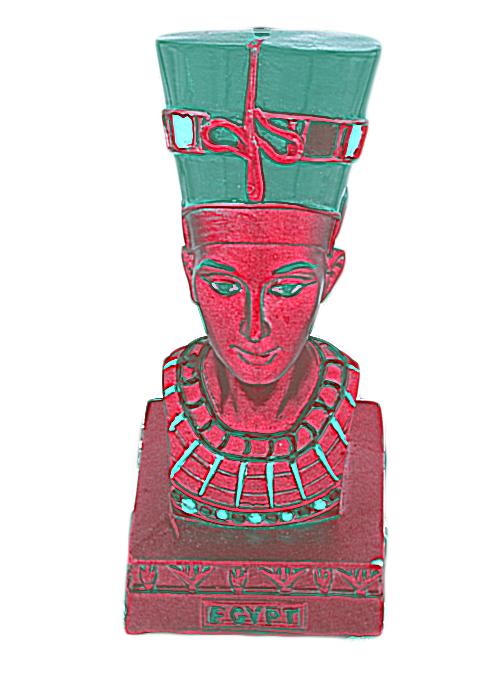}
            \includegraphics[width = 0.27\linewidth]{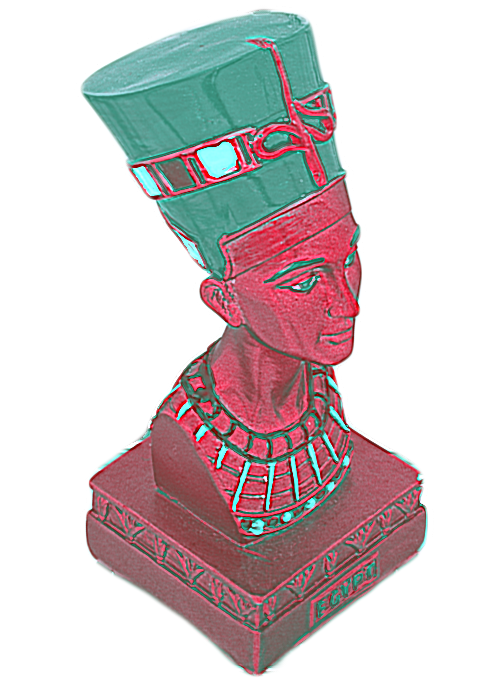}
        \end{minipage}	
        \begin{minipage}{0.03\columnwidth}	
            \centering
            \rotatebox{90}{\small A=45.8 /C=75.7}
        \end{minipage}	
    \end{minipage}	

  \caption{Impact of $\lambda$ over features. Reconstruction errors are reported on the right.}
  \label{fig:lambda}
\end{figure}

We evaluate the impact of an extensive set of factors over the final reconstruction. For all figures mentioned in this subsection, images in the first and second column are for visual comparison of learned features with a small baseline, and the third column is for large baseline comparisons. Quantitative results on 3D reconstruction are also listed on the right in vertical text. Moreover, the second row in~\figref{fig:lambda} shows the common baseline result, which is omitted in all other figures due to limited space.

In~\figref{fig:lambda}, we first study the impact of $\lambda$. When $\lambda$ is small, the loss on positive pairs dominates, leading to degenerated features that are less spatially distinctive. On the other hand, when $\lambda$ is large, the loss on negative pairs kicks in to produce high-frequency spatial features, at the cost of view stability. The current choice of $0.01$ is selected after balancing the two factors.

\begin{figure}
    \begin{minipage}{\columnwidth}
        \centering
        \begin{minipage}{0.03\columnwidth}	
            \centering
            \rotatebox{90}{\small 5x5x1}
        \end{minipage}	
        \begin{minipage}{0.92\columnwidth}	
            \centering
            \includegraphics[width = 0.27\linewidth]{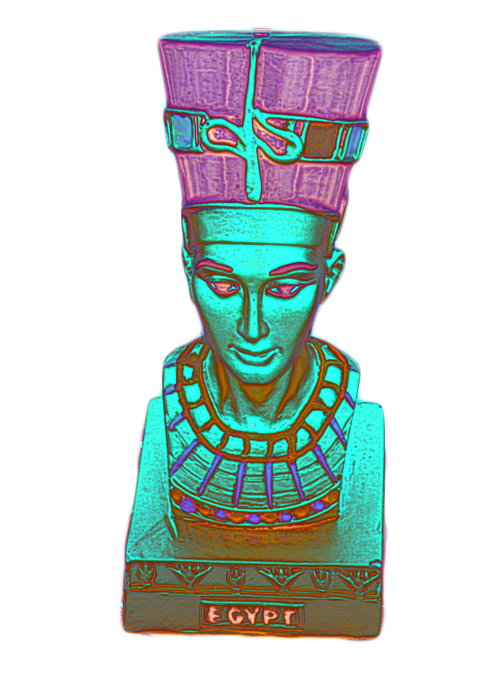}
            \includegraphics[width = 0.27\linewidth]{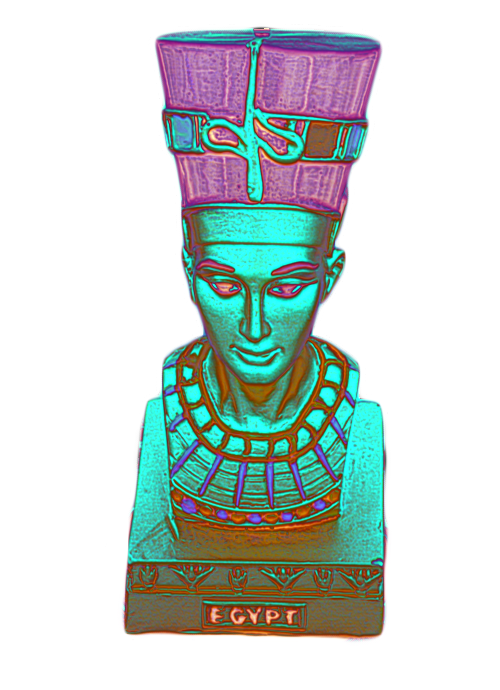}
            \includegraphics[width = 0.27\linewidth]{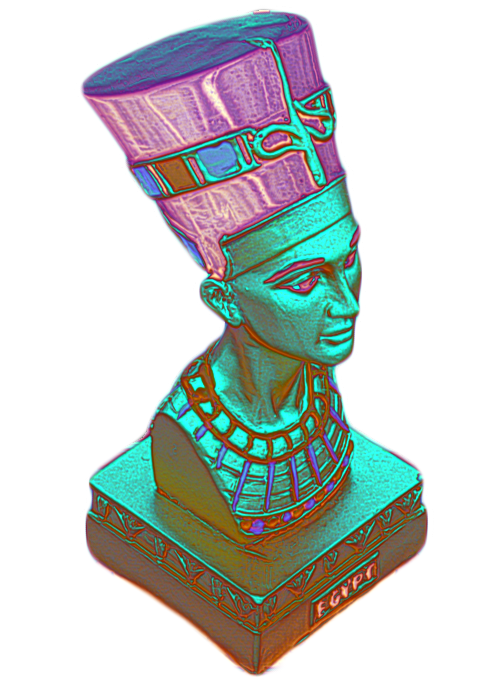}
        \end{minipage}	
        \begin{minipage}{0.03\columnwidth}	
            \centering
            \rotatebox{90}{\small A=41.0/C=71.4}
        \end{minipage}	
    \end{minipage}	

    \begin{minipage}{\columnwidth}
        \centering
        \begin{minipage}{0.03\columnwidth}	
            \centering
            \rotatebox{90}{\small 1x1x5}
        \end{minipage}	
        \begin{minipage}{0.92\columnwidth}	
            \centering
            \includegraphics[width = 0.27\linewidth]{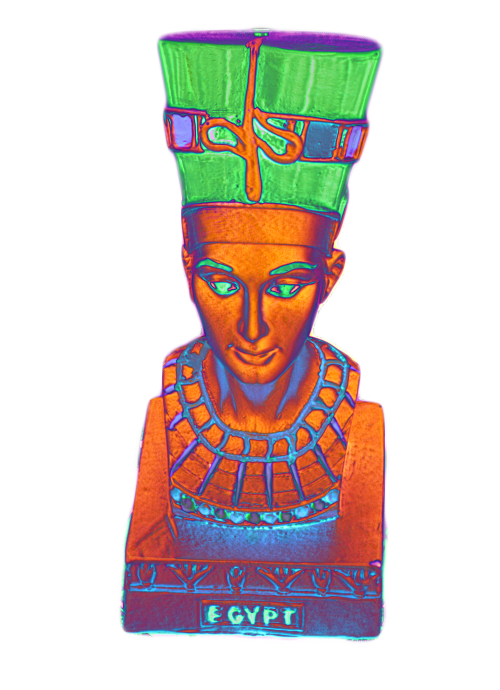}
            \includegraphics[width = 0.27\linewidth]{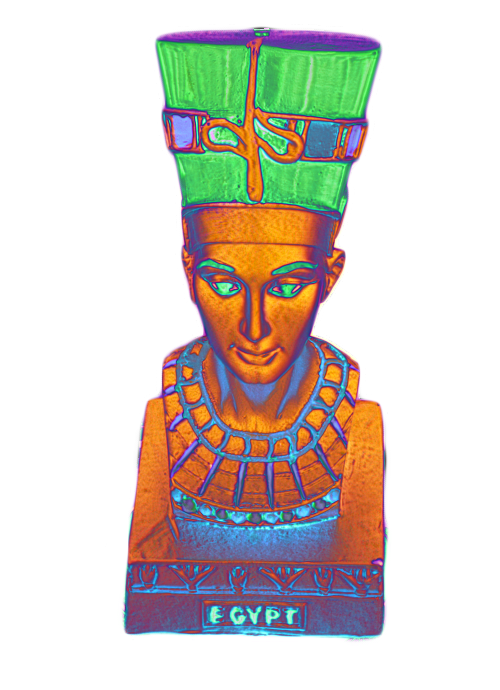}
            \includegraphics[width = 0.27\linewidth]{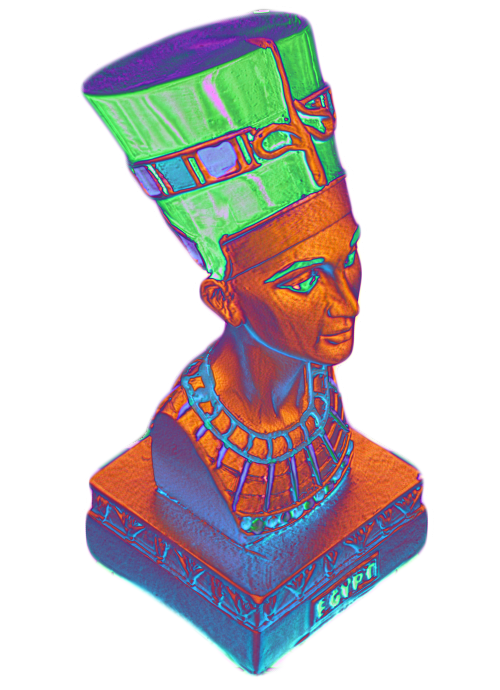}
        \end{minipage}	
        \begin{minipage}{0.03\columnwidth}	
            \centering
            \rotatebox{90}{\small A=47.0/C=77.2}
        \end{minipage}	        
    \end{minipage}
  \caption{Impact of tensor size over features. Reconstruction errors are reported on the right.}
  \label{fig:tensorsize}
\end{figure}

We test two interesting tensor sizes in~\figref{fig:tensorsize}, 5$\times$5$\times$1 and 1$\times$1$\times$5. The former discards the differential cues in the neighboring views and degenerates to a pure image-domain technique, resulting in features that are less sensitive to surface details and thus a less satisfactory shape. The latter does not take the spatial neighborhood into account and solely relies on the angular domain, producing a better reconstruction with a surprisingly limited tensor size. This demonstrates the effectiveness of cues in densely sampled views.
\begin{figure}
    \begin{minipage}{\columnwidth}
        \centering
        \begin{minipage}{0.03\columnwidth}	
            \centering
            \rotatebox{90}{\small Fixed 4-point Lighting}
        \end{minipage}	
        \begin{minipage}{0.92\columnwidth}	
            \centering
            \includegraphics[width = 0.27\linewidth]{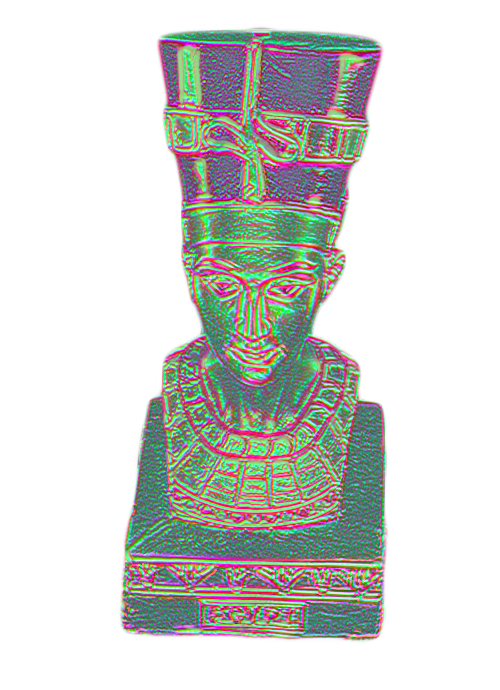}
            \includegraphics[width = 0.27\linewidth]{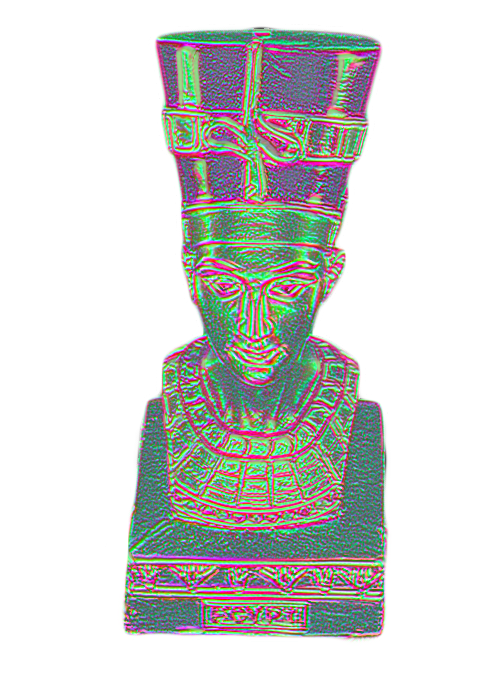}
            \includegraphics[width = 0.27\linewidth]{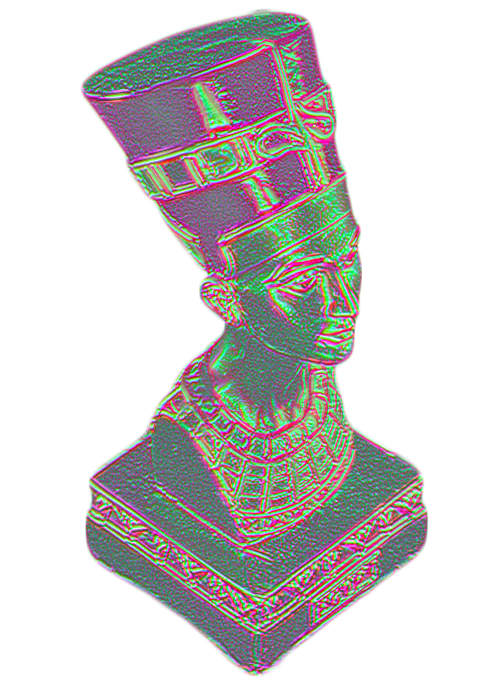}
        \end{minipage}	
        \begin{minipage}{0.03\columnwidth}	
            \centering
            \rotatebox{90}{\small A=40.5/C=58.5}
        \end{minipage}	
    \end{minipage}	
  \caption{Impact of illumination over features. Reconstruction errors are reported on the right.}
  \label{fig:lightingnum}
\end{figure}

\begin{figure}
    \begin{minipage}{\columnwidth}
        \centering
        \begin{minipage}{0.03\columnwidth}	
            \centering
            \rotatebox{90}{\small 7x7}
        \end{minipage}	
        \begin{minipage}{0.92\columnwidth}	
            \centering
            \includegraphics[width = 0.27\linewidth]{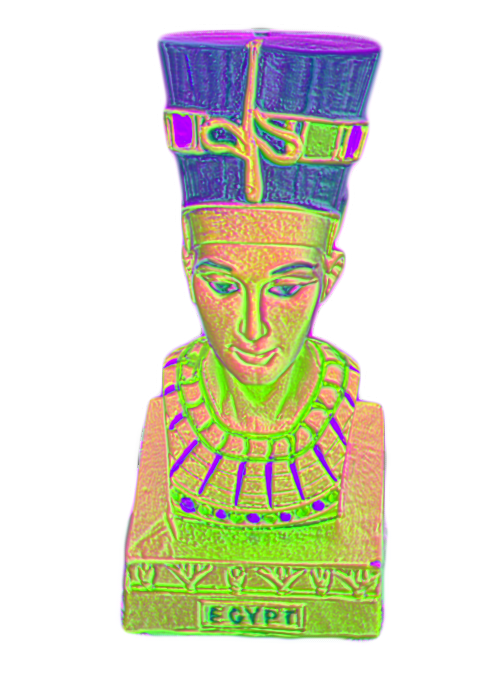}
            \includegraphics[width = 0.27\linewidth]{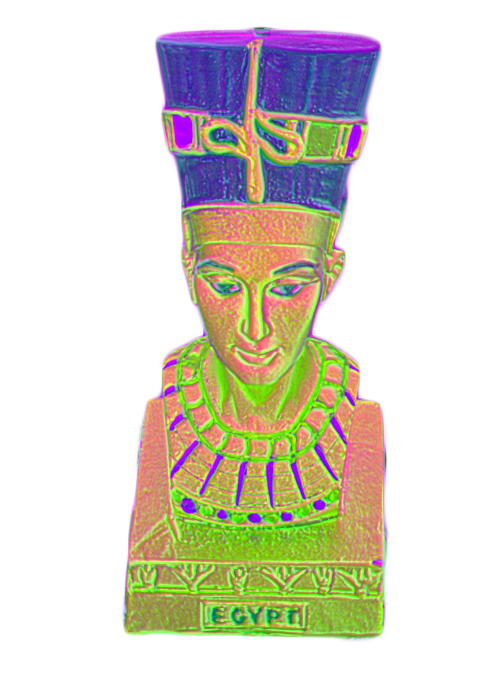}
            \includegraphics[width = 0.27\linewidth]{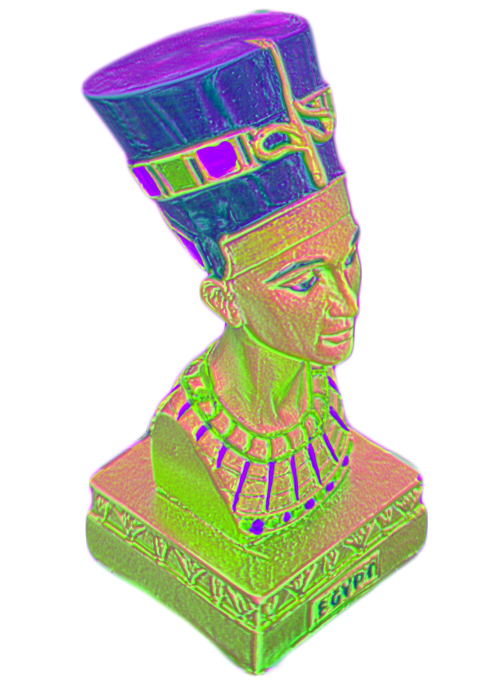}
        \end{minipage}	
        \begin{minipage}{0.03\columnwidth}	
            \centering
            \rotatebox{90}{\small A=47.3/C=76.4}
        \end{minipage}	
    \end{minipage}	
    \begin{minipage}{\columnwidth}
        \centering
        \begin{minipage}{0.03\columnwidth}	
            \centering
            \rotatebox{90}{\small 3x3}
        \end{minipage}	
        \begin{minipage}{0.92\columnwidth}	
            \centering
            \includegraphics[width = 0.27\linewidth]{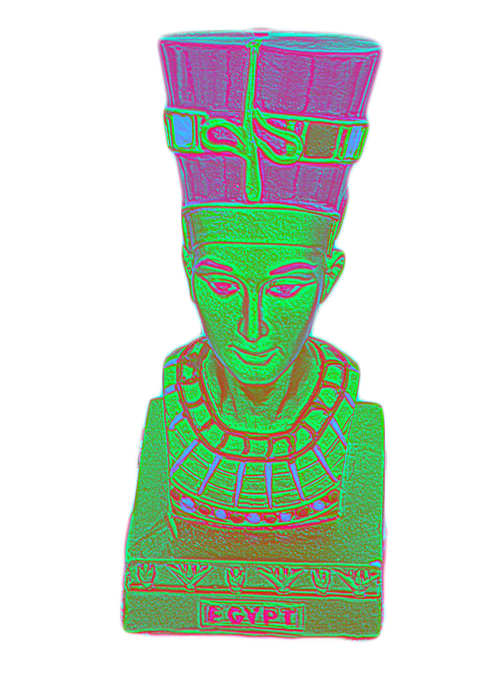}
            \includegraphics[width = 0.27\linewidth]{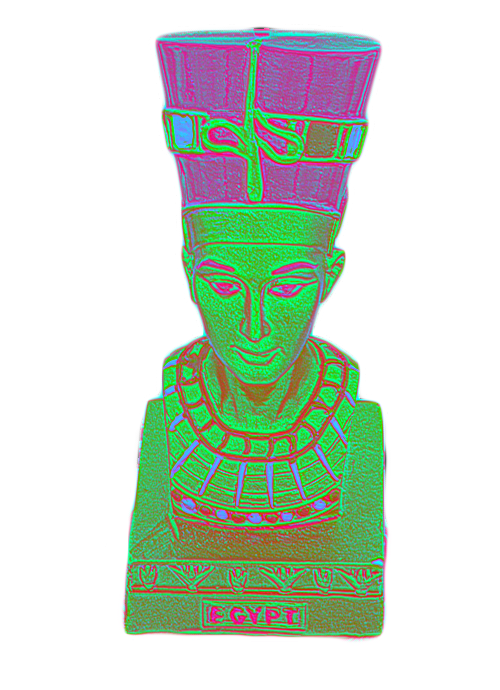}
            \includegraphics[width = 0.27\linewidth]{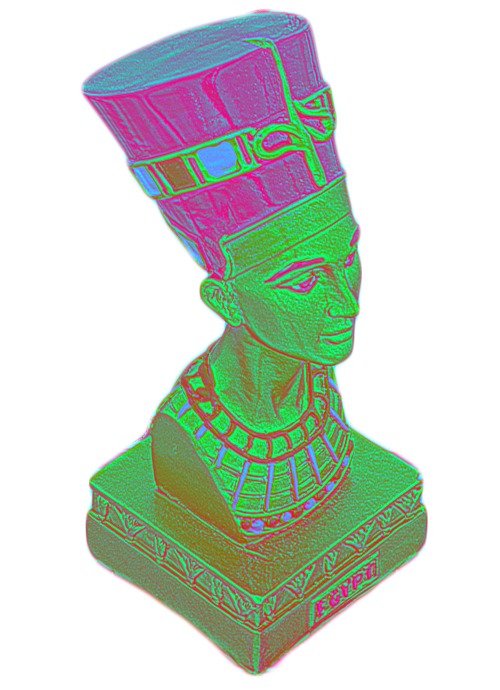}
        \end{minipage}	
        \begin{minipage}{0.03\columnwidth}	
            \centering
            \rotatebox{90}{\small A=47.6/C=78.8}
        \end{minipage}	
    \end{minipage}	
  \caption{Impact of neighborhood size ($w_{\operatorname{neg}} \times w_{\operatorname{neg}}$) of negative samples over features. Reconstruction errors are reported on the right.}
  \label{fig:neighborhoodsize}
\end{figure}
\begin{figure}
    \begin{minipage}{\columnwidth}
        \centering
        \begin{minipage}{0.03\columnwidth}	
            \centering
            \rotatebox{90}{\small Image\# = 180}
        \end{minipage}	
        \begin{minipage}{0.92\columnwidth}	
            \centering
            \includegraphics[width = 0.27\linewidth]{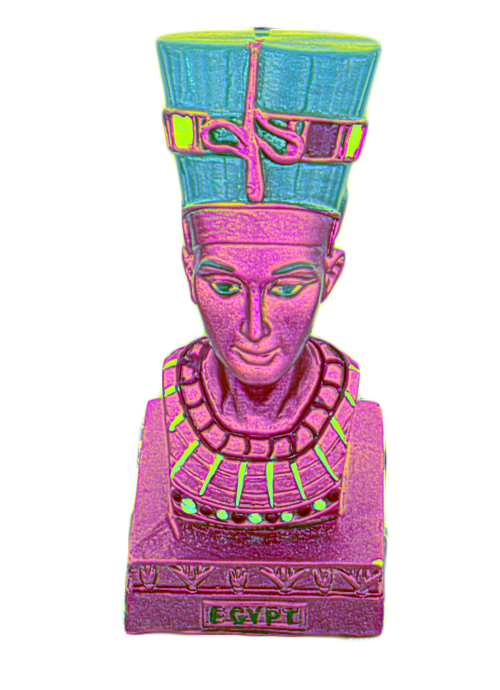}
            \includegraphics[width = 0.27\linewidth]{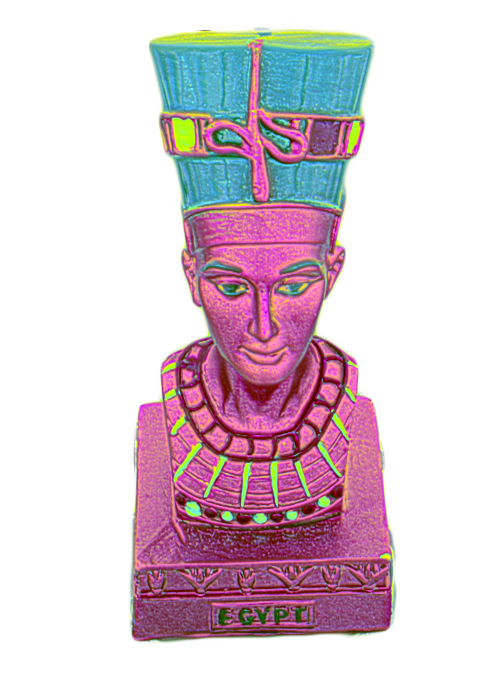}
            \includegraphics[width = 0.27\linewidth]{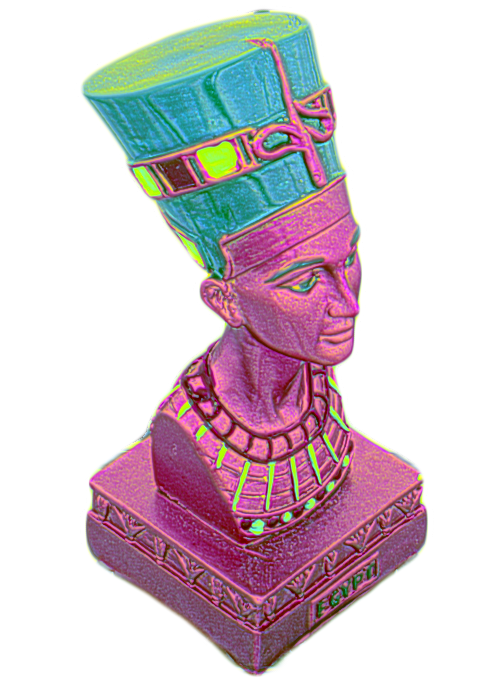}
        \end{minipage}	
        \begin{minipage}{0.03\columnwidth}	
            \centering
            \rotatebox{90}{\small A=51.5/C=63.4}
        \end{minipage}	
    \end{minipage}	

    \begin{minipage}{\columnwidth}
        \centering
        \begin{minipage}{0.03\columnwidth}	
            \centering
            \rotatebox{90}{\small Image\# = 90}
        \end{minipage}	
        \begin{minipage}{0.92\columnwidth}	
            \centering
            \includegraphics[width = 0.27\linewidth]{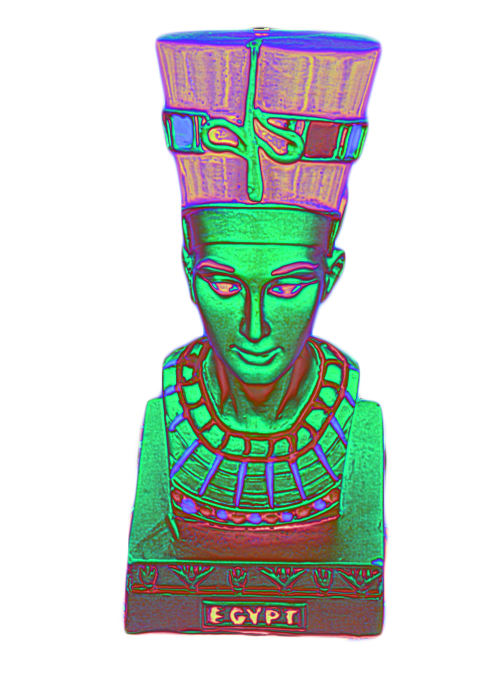}
            \includegraphics[width = 0.27\linewidth]{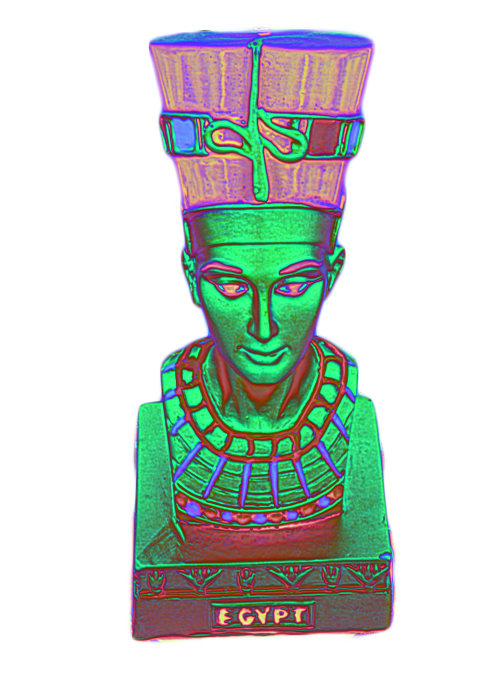}
            \includegraphics[width = 0.27\linewidth]{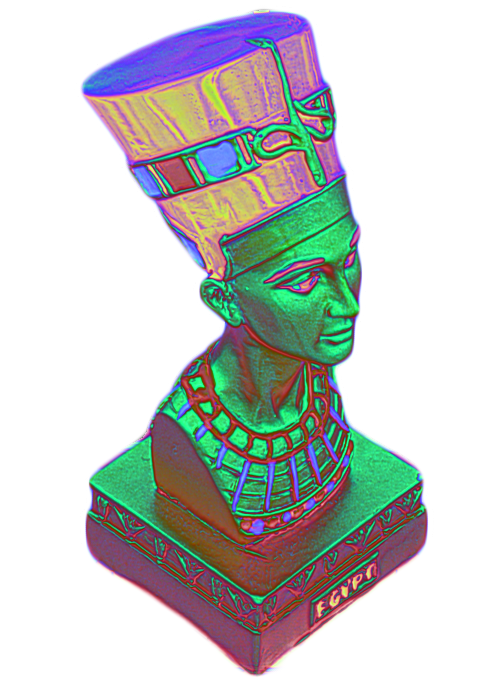}
        \end{minipage}	
        \begin{minipage}{0.03\columnwidth}	
            \centering
            \rotatebox{90}{\small A=51.3/C=40.1}
        \end{minipage}	
    \end{minipage}	
  \caption{Impact of angular sampling rate over features. Reconstruction errors are reported on the right.}
  \label{fig:rotationspeed}
\end{figure}

In~\figref{fig:neighborhoodsize}, we further investigate the neighborhood size used in negative pair sampling (\sec{sec:loss}). Due to effective spatial propagation, different neighborhood sizes end up with similar results. In~\figref{fig:rotationspeed}, we experiment with different angular sampling rates. While a bigger angular interval ($2^{\circ}$/$4^{\circ}$) corresponds to a larger receptive field of the tensor, the coherence (i.e., differential structures) among neighboring views is reduced and the number of captured views as well. These two factors lead to less complete results.

Finally, the framework is extended to a fixed 4-point lighting pattern (~\figref{fig:lightingnum}). As expected, the quality is not as good as the baseline, due to fewer degrees of freedom in the optimization. Nevertheless, the experiment demonstrates the flexibility of our framework to adapt to different configurations. It will be interesting to extend to more complex illumination conditions, such as multiple patterns at each view and alternating patterns that change with the view, to further exploit the rich photometric information in these cases.

\subsection{Additional Applications}
\label{sec:app}
\begin{figure}
    \centering
    \begin{minipage}{0.32\linewidth}	
        \centering
        \small Photo
    \end{minipage}	
    \begin{minipage}{0.32\linewidth}	
        \centering
        \small DiFT
    \end{minipage}	
    \begin{minipage}{0.32\linewidth}	
        \centering
        \small Laplacian
    \end{minipage}	
    
    \includegraphics[width = 0.32\linewidth]{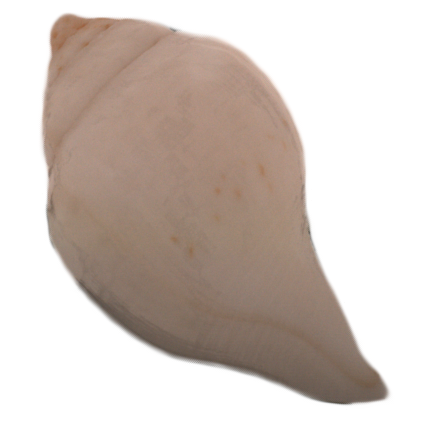}
    \includegraphics[width = 0.32\linewidth]{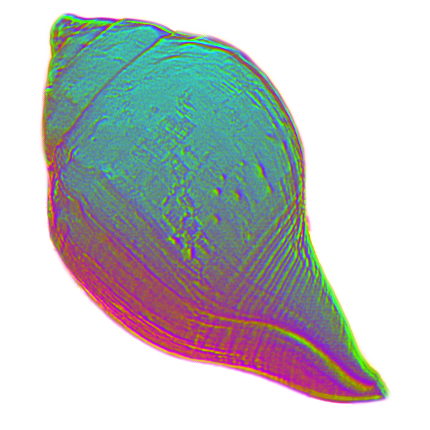}
    \includegraphics[width = 0.32\linewidth]{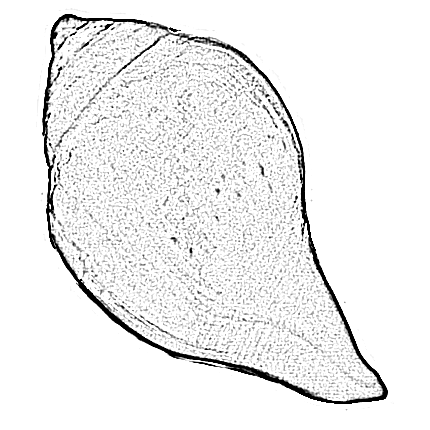}
  
  \caption{Comparison between DiFT and the image-domain Laplacian. Our features better bring out intricate geometric details. Please zoom-in on a computer screen for a better visualization.}
  \label{fig:crack}
\end{figure}
Here we briefly describe two additional applications. First, despite the presence of complex appearance, DiFT features visualize/magnify the intricate geometric details using as few as 5 input images at neighboring views (\figref{fig:crack}). Such details might be difficult to spot in the original photograph, or even after applying a conventional image-domain Laplacian filter. As shown in the figure, our features bring out the natural growth pattern on the surface of the conch, which might be helpful in fields like biology or archaeology.

\begin{figure}
    \centering
    \begin{minipage}{0.32\linewidth}	
        \centering
        \small Photo
    \end{minipage}	
    \begin{minipage}{0.32\linewidth}	
        \centering
        \small Style\#1
    \end{minipage}	
    \begin{minipage}{0.32\linewidth}	
        \centering
        \small Style\#2
    \end{minipage}	
    
    \includegraphics[width = 0.32\linewidth]{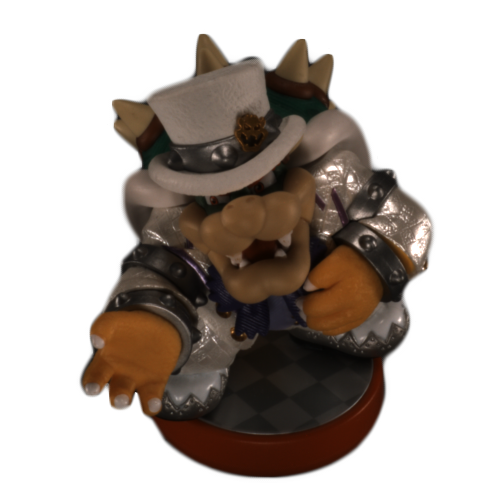}
    \includegraphics[width = 0.32\linewidth]{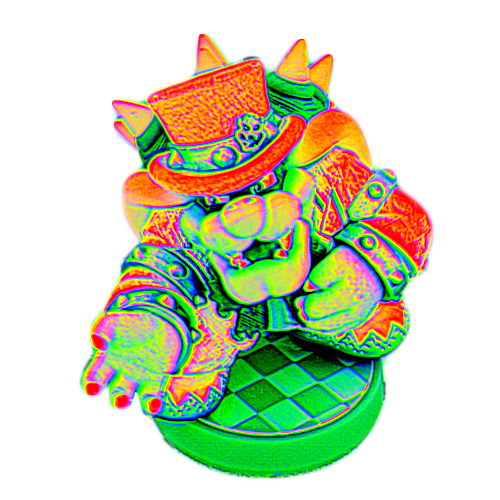}
    \includegraphics[width = 0.32\linewidth]{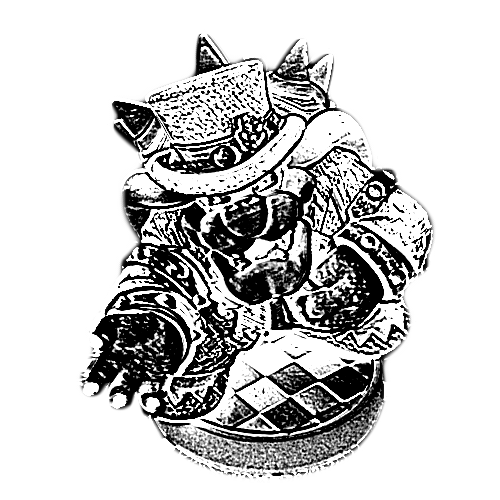}
  \caption{Cartoon-/sketch-like stylization with our features.}
  \label{fig:cartoon}
\end{figure}

In~\figref{fig:cartoon}, we convert the feature maps with simple per-pixel operations into cartoon-/sketch-like stylization results that are stable across different views. DiFT may offer a unique perspective to computational stylization in the presence of complex appearance~\cite{bousseau2013gloss}.

\section{Limitations and Future Work}
Our work is subject to a few limitations. First, the training data depend on the availability of high-quality digitized objects with separate representations of shape and appearance, a key resource that is still lacking nowadays. In addition, global illumination is not considered in training tensor generation, though more complete simulations can be performed at the expense of a substantially increased computation burden. Finally, the current framework supports reflectance only.

For future work, it will be promising to extend DiFT to handle 1D/2D translational motions, or even irregular motions, as long as the view specification for each image can be accurately calibrated and reparameterized to be used with deep learning. It will also be interesting to reconstruct the 6D appearance from differential cues in conjunction with the geometry. Following~\sec{sec:app}, we hope that DiFT will serve as a fundamental low-level feature descriptor that might find useful applications in a variety of fields beyond computer graphics.


\bibliographystyle{ACM-Reference-Format}
\bibliography{diffdiff}


\end{document}